\documentclass[10pt,twocolumn,letterpaper]{article}

\usepackage{iccv}
\usepackage{times}
\usepackage{epsfig}
\usepackage{graphicx}
\usepackage{amsmath}
\usepackage{amssymb}

\usepackage{amsfonts}
\newcommand\norm[1]{\left\lVert#1\right\rVert}

\newcommand{\I}{\mathcal{I}}
\newcommand{\F}{\mathcal{F}}
\usepackage{subcaption}
\usepackage{float}
\usepackage{adjustbox}
\usepackage{booktabs}

\usepackage[pagebackref=true,breaklinks=true,letterpaper=true,colorlinks,bookmarks=false]{hyperref}

\iccvfinalcopy 

\ificcvfinal\fi
\begin{document}

\title{Image Resizing by Reconstruction from Deep Features}

\author{Moab Arar, Dov Danon, and Daniel Cohen-Or\\
Tel Aviv University
\and
Ariel Shamir\\
The Interdisciplinary Center  
}

\maketitle
\begin{abstract}

Traditional image resizing methods usually work in pixel space and use various saliency measures.  
The challenge is to adjust the image shape while trying to preserve important content. 
In this paper we perform image resizing in feature space where the deep layers of a neural network contain rich important semantic information.
We directly adjust the image feature maps, extracted from a pre-trained classification network, and reconstruct the resized image using a neural-network based optimization.
This novel approach leverages the hierarchical encoding of the network, and in particular, the high-level discriminative power of its deeper layers, that recognizes semantic objects and regions and allows maintaining their aspect ratio.
Our use of reconstruction from deep features diminishes the artifacts introduced by image-space resizing operators. We evaluate our method on benchmarks, compare to alternative approaches, and demonstrate its strength on challenging images.

\end{abstract}

\section{Introduction}
\begin{figure}
\centering
\begin{subfigure}{0.2\textwidth}
\centering
\includegraphics[height=3.6cm]{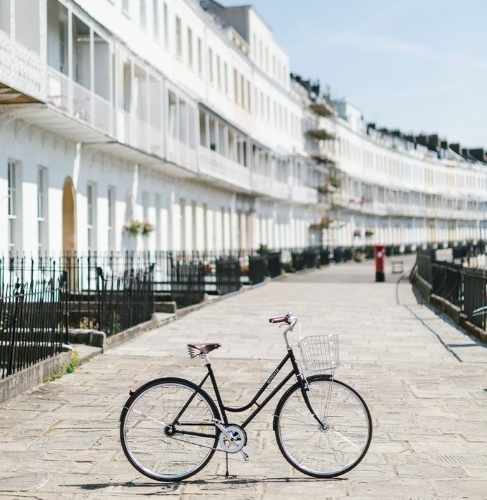}
\caption{}
\end{subfigure}
\begin{subfigure}{0.07\textwidth}
\centering
\includegraphics[height=3.6cm]{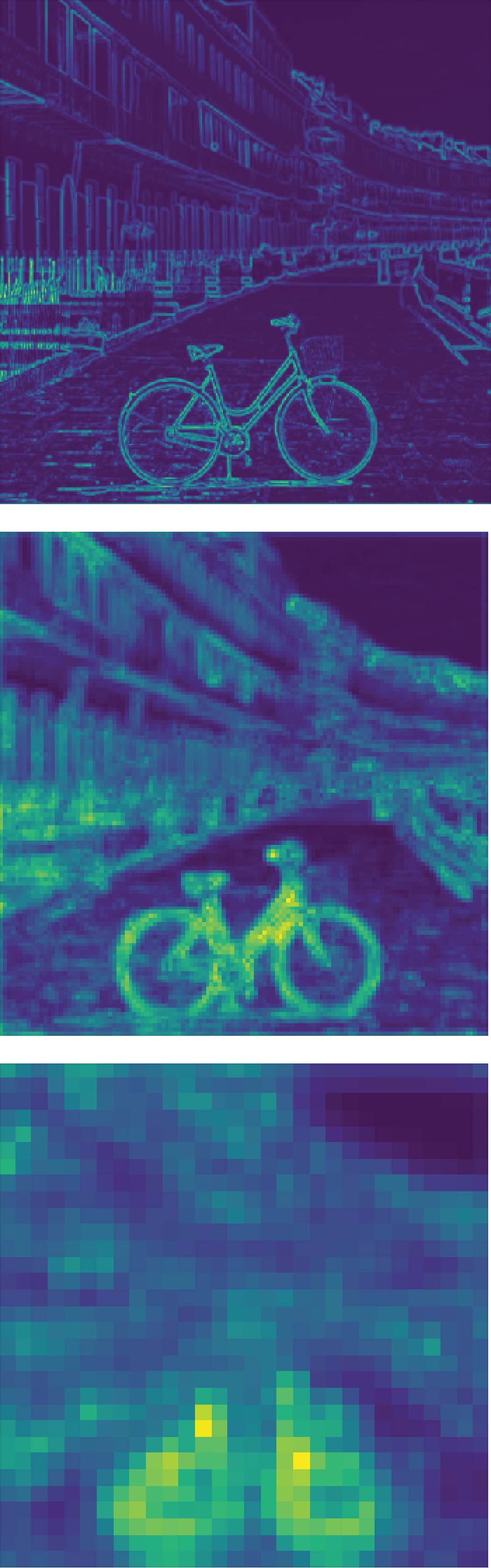}
\caption{}
\end{subfigure}
\begin{subfigure}{0.12\textwidth}
\centering
\includegraphics[height=3.6cm]{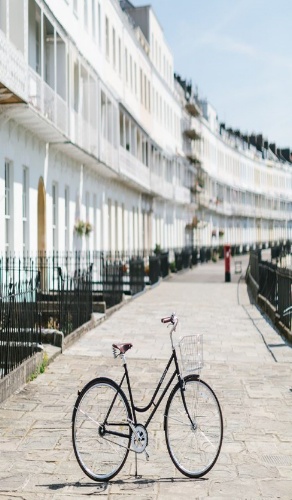}
\caption{}
\end{subfigure}

\begin{subfigure}{0.13333\textwidth}
\centering
\includegraphics[height=3.6cm]{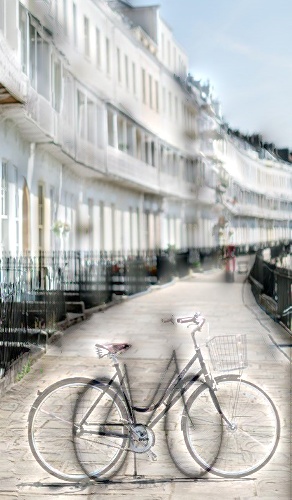}
\caption{}
\end{subfigure}
\begin{subfigure}{0.13333\textwidth}
\centering
\includegraphics[height=3.6cm]{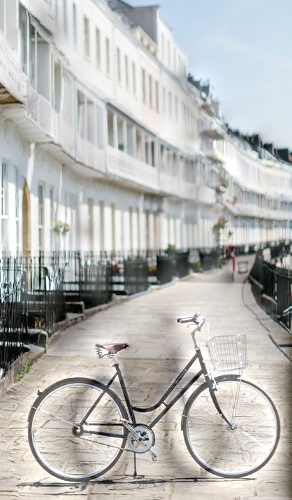}
\caption{}
\end{subfigure}
\begin{subfigure}{0.13333\textwidth}
\centering
\includegraphics[height=3.6cm]{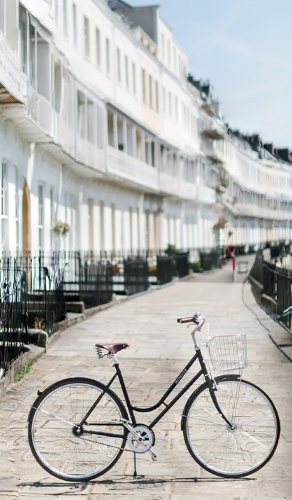}
\caption{}
\end{subfigure}
\caption{Given an input image (a), our deep network resizing method first adjusts the size of the feature maps of a deep neural network (b), while protecting important semantic regions, and then reconstructs a retargeted image using iterative optimization (c-f). Note how starting from a linear scaled image (c), the iterations manage to reconstruct the shape of the bicycle (d-f), which is the main semantic object in the image, while minimizing artifacts.}
\label{fig:teaser}
\end{figure}
The media resizing problem had been widely studied in the last decade and many content-aware methods have been developed~\cite{Avidan07, Wolf07, Rubinstein08, Wang08, Pritch09, Guo09, Krahenbuhl09, Rubinstein09, Wu2010, Panozzo12,Cho17,Shocher2018}. The main objective of these methods is to change the size of the input while maintaining the appearance of important regions such as salient object, and reducing visual artifacts. These two objectives can be seen as two quality measures that are sometimes contradicting. The first one measures how semantically close the resulting image is to the original one by preserving its important parts, and the second one measures the resemblance of the result to a natural image by reducing artifacts (see~\cite{BlauMichaeli18}).

Most techniques first employ some saliency measurement techniques to decide which regions of the image are more important. Then, an image resizing operator is used to create the resized image while preserving these regions, and hoping to introduce less artifacts. Both of these steps are still challenging. First, common saliency measures account for low-level features only, while disregarding important high-level semantics. Second, current resizing operators do not directly account for the second quality measure of maintaining the natural look of the resulting image.

In this work we present Deep Network Resizing (DNR) as a method that deals with the two aforementioned challenges using neural networks.
First, we exploit the ability of pre-trained networks to analyze and encode both low-level and high-level features to identify important parts in the image. In addition, we employ a back-propagation aided optimization method to directly preserve both the structure of important regions and the natural image appearance of the result. This reduces many of the artifacts issues which exist in traditional approaches, and integrates analysis and synthesis based on neural networks in an image resizing technique.

The key idea of DNR is that instead of applying image resizing operators on the pixels of the image, it applies them in feature space, on the feature maps of deep layers of a pre-trained Convolutional Neural Network (Figure~\ref{fig:method_sketch}). This allows content removal to concentrate on regions of the image that are semantically irrelevant. We show that DNR discards insignificant data, which in turn, preserves the semantic encoding of the input image. The operator we demonstrate our approach with is seam carving algorithm~\cite{Avidan07}. 

After the image is reconstructed using back-propagation based optimization we perform a refinement step. In this step, a grid-sampler layer is used, allowing only a change in the mapping of pixels while optimizing for the same objective. This step increases the natural look of the resulting images, by further reduction of artifacts. 

Our main contributions are:
\vspace{-0.1cm}
\begin{itemize}
\item
Utilizing the semantic guidance of deep layers of a CNN for image importance in resizing. 
\item
Applying seam-carving in feature-space instead of image-space.
\item
Reducing artifacts of reconstructed images by optimization using grid-sampling.
\item
Presenting Deep Network Resizing, a method for image resizing using neural networks.
\end{itemize}

\section{Related Work}

\textbf{Image Processing Techniques:}
A considerable work on content-aware media retargeting have been carried out in the field of image processing, and it is widely common to classify it into \textit{discrete methods}~\cite{Avidan07, Rubinstein08, Pritch09, Rubinstein09} and \textit{continuous methods}~\cite{Wolf07, Wang08, Krahenbuhl09, Guo09,Wu2010,Panozzo12} (refer to~\cite{Kiess18, Vaquero15} for comprehensive coverage on content-aware retargeting).

\textit{Discrete methods: } Seam carving was introduced by Avidan \etal~\cite{Avidan07}, which performs retargeting by repeatedly inserting or removing connected paths of pixels (called seams), passing through low importance regions. Later, Rubinstein \etal~\cite{Rubinstein08} improved seam carving using a look-forward energy map, which measures the amount of energy introduced on seam removal or insertion. Pritch \etal~\cite{Pritch09} introduced shift-maps for pixels re-arrangement, and formulated graph-labeling problem for various image editing applications, including retargeting. Rubinstein \etal~\cite{Rubinstein09} combined different retargeting operators through finding paths in a multidimensional space which dictates sequence of retargeting operations on the input media.  

\textit{Continuous  methods: }Wolf \etal~\cite{Wolf07} introduced a map that is determined by three importance measures in order to devise a system of linear equations that defines a mapping of source pixels into their corresponding location in the target image. Wang \etal~\cite{Wang08} compute a deformed mesh-grid by assigning a scale factor for each quad in the grid. They proposed two penalties, that encourages their solution to linear-scale quads of high-importance and allow higher deformation on low-importance grids. Krahenbuhl \etal~\cite{Krahenbuhl09} uses an energy map, consisting of many automatic constraints and user defined constraints on key frames, in order to compute a non-uniform pixel accurate warping on video streams.  Guo \etal~\cite{Guo09} define a saliency-based triangles mesh representation, and use constrained mesh parametrization problem to compute the retargeting solution. Wu \etal \cite{Wu2010} detect symmetric parts in the image and then carry summarization operation on the symmetric regions and warping on non-symmetric parts. Panoozzo \etal~\cite{Panozzo12} use axis-aligned representation which overcomes the optimization's complexity when using 2D parametric representation of mesh deformation. The authors~\cite{Panozzo12} later find the deformation parameters by solving a simple quadratic problem with linear constraints.  

\textbf{Deep Learning Techniques:} The superiority of CNNs in solving computer vision tasks, including Image Recognition~\cite{Krizhevsky12, Simonyan14, Szegedy15, He16}, Segmentation and Detection~\cite{Girshick14, Girshick15, HeZR014,RenHGS15}, has already been established over the past years.

One possible approach of using deep learning to solve the retargeting problem would be to gather a training set of original and retargeted images and use supervised learning. However, it is very difficult to gather such a set as each image must support numerous retargeting sizes and there is no ground-truth method to apply this. Moreover, using manual retargeting may produce different results for different artists.

Cho \etal~\cite{Cho17} have proposed a weakly and self supervised learning for image retargeting. The authors use the semantic encoding of pre-trained networks and devise a decoder that produces an attention map. The attention map is then combined with a shift-layer in order to obtain the target image. Unlike DNR, the authors~\cite{Cho17} train their network on a given dataset, where the objective is to minimize structural damages while maintaining the detection score of the image, as given by the pre-trained CNN. DNR, however performs per-input analysis, and presents a solution that uses the strengths of deep learning in understanding the semantics of the image and in correcting images. DNR utilizes different retargeting operators to produce a feature representation of the target image (see comparison in Figure~\ref{fig:ours_vs_wssdcnn}). In a more recent paper, Shocher \etal~\cite{Shocher2018} proposed a Generative Adverserial Network (GAN) method for synthesizing images that can be considered a type of retargeting. The authors learn the patch distribution of the input image, and use this to generate images with similar patch statistics as the input image. However, the resulting image can have a different structure than the original image and still contain some artifacts. 

An independent work was recently proposed in~\cite{DBLP:journals/corr/abs-1811-07793}, in which the authors also perform retargeting in feature space. However, there are two fundamental differences between their work and ours. First, they preform retargeting by sampling columns of deep feature maps at a constant rate, where we combine several deep retargeting operators. Further, the authors adapt methods in~\cite{Liao:2017:VAT:3072959.3073683} and perform warping on the input image using PatchMach~\cite{Barnes:2009:PRC:1531326.1531330}, where our image is reconstructed via pure synthesis procedure (see comparison in Figure~\ref{fig:ours_vs_deepir}). 
\section{Method}
\label{sec:method}
\begin{figure}
    \centering
    \includegraphics[]{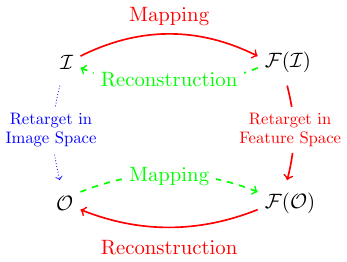}
    \caption{Conventional resizing approaches act in image space (blue arrow), while our deep-resizing approach (red) applies resizing in the semantic feature space. We map image $\mathcal{I}$ to feature maps $\mathcal{F(I)}$ in feature space using a CNN. Then, we resize in feature space to create $\mathcal{F(O)}$. Lastly, we use back propagation optimization to reconstruct $\mathcal{O}$. In other words, instead of reconstructing the original image $\mathcal{I}$ from $\mathcal{F(I)}$ (green arrow), we reconstruct the resized image $\mathcal{O}$ from $\mathcal{F(O)}$, which is the hypothetical mapping of $\mathcal{O}$ to feature space.
    }
    \label{fig:method_sketch}
\end{figure}

Conventional image resizing apply pixel-manipulations on the image. In this work, we propose a new approach, where resizing is applied in feature space, and the results are mapped back into image space by reconstruction (see Figure~\ref{fig:method_sketch}). 
Our key idea is leveraging deep features of a pre-trained CNN, which encode valuable latent semantics. By applying the resizing operators in feature-space we create \emph{target feature maps}, where semantic information is kept unharmed. To reconstruct the output image we use optimization that iteratively minimizes the difference between the target feature maps and the actual feature maps of the optimized image. 

Let $\I$ be an input image of size $(h,w)$. Assume we use a pre-trained deep-network with $L$ layers, we define the activation values of all neurons in level $i$ applied on input $\I$ as the $i$-th \emph{feature map} $\F_i(\I)$. $\F(\I)$ is the set of all feature maps for $1 \leq i \leq L$:
$$ \F(\I) = \{ \F_1(\I),\F_2(\I),\dots{},\F_L(\I)\}.$$
Each feature map $\F_{i}(\I)$ has a certain number of channels, and a spatial dimension that depends on the size of the input $\I$.
We denote by $(h_{i}^{\I},w_{i}^{\I},c_{i})$ the height, width, and number of channels of the $i$-th feature map. 

Given the target size $(h',w')$, the task of resizing in image-space is to obtain an image $\mathcal{O}$ of size $(h',w')$, while maintaining important regions in $\I$ and reducing artifacts as much as possible. 
The resizing task in feature-space is defined as obtaining a set of \emph{target feature maps}: 
$$\F' = \{ \F'_1,\F'_2,\dots{},\F'_L\}$$
such that for each level $i$, $\F'_i$ matches the dimension of the $i$-th feature map $\F_i(\mathcal{O})$ of the resized image $\mathcal{O}$, while preserving the important regions of the original image's feature maps $\F_{i}(\I)$. That is, the dimensions of $\F'_i$ are $(h_{i}^{\mathcal{O}},w_{i}^{\mathcal{O}},c_{i})$ but it contains the most important information in $\F_{i}(\I)$.

To obtain the actual resized image $\mathcal{O}$ we assume that $\F' = \F(\mathcal{O})$, the hypothetical mapping of $\mathcal{O}$ to feature space, and reconstruct $\mathcal{O}$ by minimizing the difference between the output feature maps and the target feature-maps using back-propagation. Since important regions in various levels are maintained in the target feature maps $\F'$, the reconstructed image $\mathcal{O}$ preserves them as well. Lastly, to maintain the target image natural appearance and reduce artifacts, we apply a grid-sampler~\cite{jaderberg2015spatial} that further optimizes the constructed image.

\begin{figure}
    \centering
    \centerline{\includegraphics[width=8.5cm]{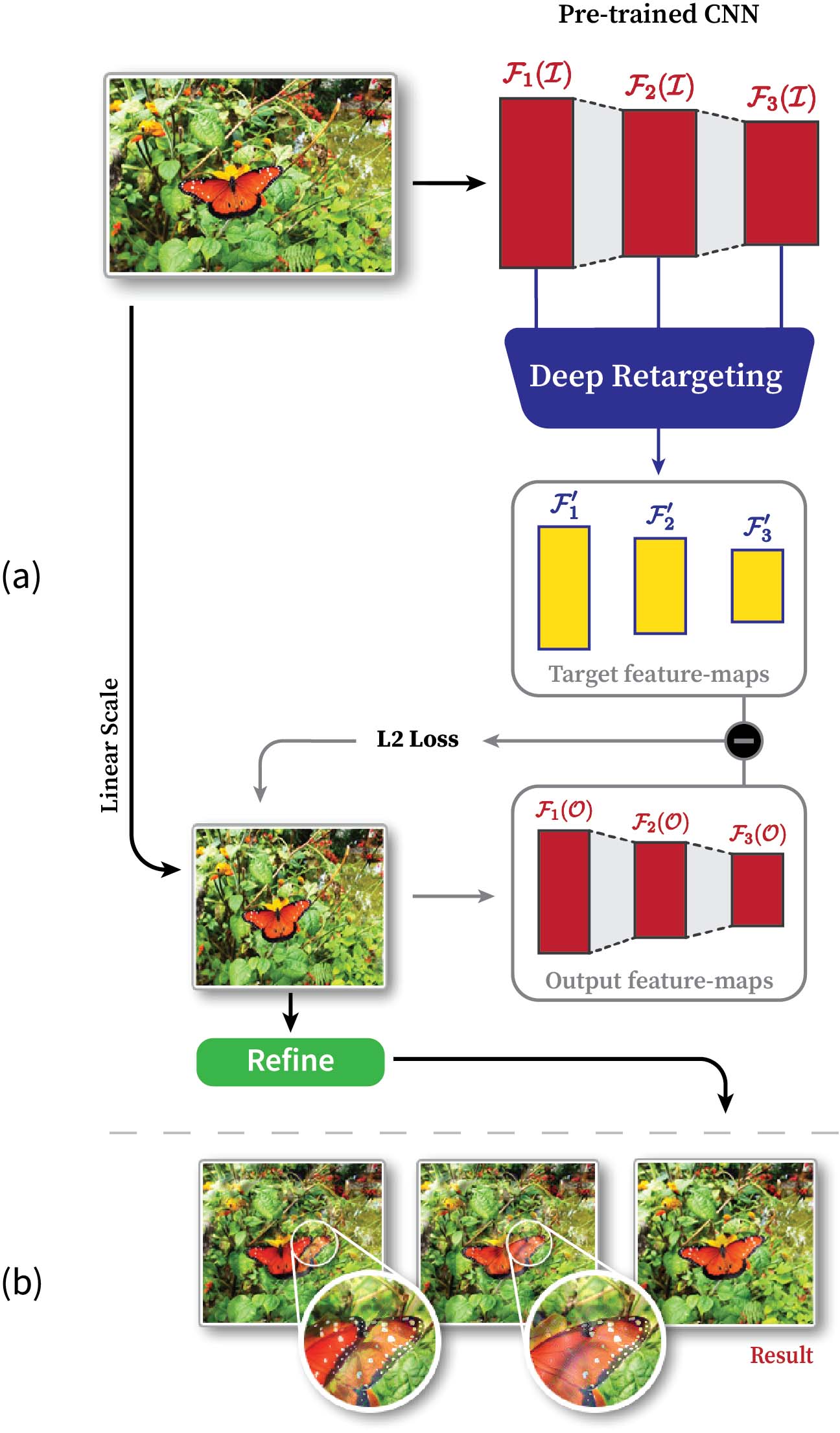}}
    \caption{Deep Network Retargeting overview. (a) The input image (\textit{left top}) is fed into a pre-trained CNN (\textit{right top}) and its features are extracted. Later, deep retargeting techniques are applied on the extracted features to produce the \emph{target feature-maps} (yellow). The output image synthesis (\textit{left bottom}) is achieved by iteratively minimizing the difference between the target feature maps and the actual feature maps of the output image. (b) Snapshots of the output image throughout the iterations: note how the semantic object is reconstructed.}
    \label{fig:method_high_level}
\end{figure}

An overview of DNR is illustrated in Figure~\ref{fig:method_high_level}. The input image (left top) is fed into a pre-trained CNN (right top) and its feature maps are extracted. Applying \textit{deep resizing} operators on selected layers yields the \textit{target feature-maps}, as illustrated in yellow in the figure. The target image (left bottom) is constructed by an optimization carried out using back-propagation: the results image is iteratively fed into the CNN, and an $L_2$-loss is computed by comparing the feature-maps of the optimized-image and the target feature-maps. This loss is back-propagated through the network to alter the target image in several iterations (depicted with a series of snapshots at the bottom of Figure~\ref{fig:method_high_level}). Lastly, to maintain the target image natural appearance and reduce artifacts, we apply a grid-sampler~\cite{jaderberg2015spatial} that further optimizes the constructed image.

In the following, we only discuss narrowing the width of the image by applying feature resizing. Similar arguments can be extended for any other target-size resizing.

\subsection{Feature Maps resizing} 
\label{sec:Featureresizing}

In our resizing method we adapt seam carving~\cite{Avidan07} and employ it to the feature maps $\F(\I)$. Guided by the feature maps $\F(\I)$, we conservatively utilize seam-carving while avoiding semantic regions. Doing so may lead to partially retargeted image, therfore, we also perform a final resizing step on the reconstructed image using grid-warping~\cite{Wolf07}. This combination allows us to harness the capabilities of the two operators: seam-carving enables the removal of homogeneous unimportant regions, and grid-warping deforms regions proportional to their importance.   

\paragraph{\textbf{Deep Seam Carving}.} 
\label{sec:DeepSeamCarving}

Seam-carving in image-space finds vertical seams as minimal one-pixel wide connected-paths on some importance map of the input image. One vertical seam removal results in reducing the image's width by one pixel. Therefore, multiple vertical seams are removed to reach the desired width of the output image. 

We extend the seam-carving algorithm by defining seam-carving on a feature-map instead of an image. First, instead of removing pixels from an image we remove neurons from the CNN layer of the feature-map. Second, because a feature map contains multiple channels, we define a seam removal as removing all neurons of the chosen seam in the same spatial location for all channels of the feature map. Third, to find minimal seams on feature-maps we use a hierarchical method to define the importance-map of each layer. Starting from the deepest, smallest in resolution, level which contains high-level semantic information, we move to shallower, larger resolution, layers that contain low-level features and refine the seams from previous layers consistently.

\begin{figure}
  \centering
 \includegraphics[height=1.5cm]{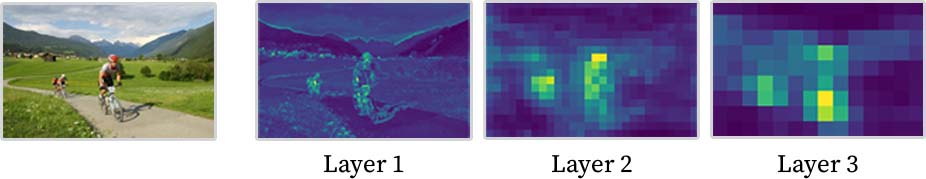}
\caption{The basic importance maps of Eq.~\ref{eq:L2SalMap} in different layers of the network. Yellow regions indicate high importance, while blue ones indicate low importance.}
\label{fig:image_with_sal_map}
\end{figure}

The basic importance-map of layer $l$ at position $(i,j)$ is defined as the $L2$-norm of the activation of the neurons along the channel axis:
\begin{equation}
\label{eq:L2SalMap}
S_{l}(i,j) = \norm{\F_l(\I)(i,j,*)}_2,
\end{equation}
where $*$ denotes all values along the channel axis (see Figure~\ref{fig:image_with_sal_map}).

\begin{figure}
\centering
\begin{subfigure}{\linewidth}
  \centering
  \includegraphics[width=8cm]{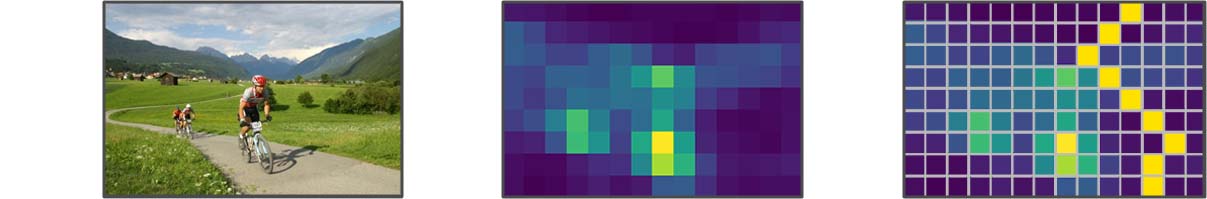}
  \caption{Deep importance map and minimal seam (layer $l$).}
  \label{fig:deep_seam_carving_explain_a}
\end{subfigure}%

\vspace{0.25cm}

\begin{subfigure}{\linewidth}
  \centering
  \includegraphics[width=8cm]{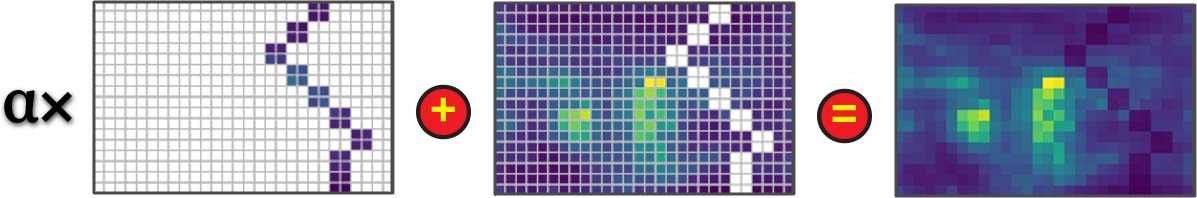}
  \caption{Reducing the importance map as defined in Eq.~\ref{eq:ModifiedL2Saliency} (for layer $l-1$) in regions where the seams pass.}
  \label{fig:deep_seam_carving_explain_b}
\end{subfigure}%

\vspace{0.25cm}

\begin{subfigure}{.6\linewidth}
  \centering
  \includegraphics[width=2.8cm]{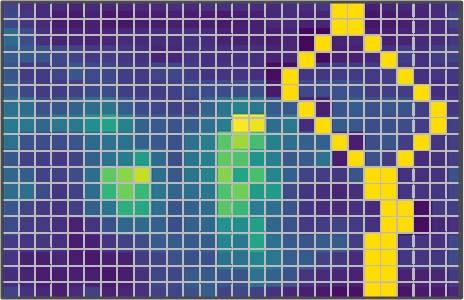}
  \caption{Chosen seams at level $l-1$ using the modified importance.}
  \label{fig:deep_seam_carving_explain_c}
  
\end{subfigure}%
\caption{Deep seam-carving applied hierarchically from layer $l$ to $l-1$. 
}
\label{fig:deep_seam_carving_explain}
\end{figure}

We start by applying seam-carving on the deepest layer $L$ in the hierarchy using the importance-map defined in Equation~\ref{eq:L2SalMap}. This map is useful since deep-layer neurons have higher activation in semantic regions. As we move up the hierarchy from level $l$ to level $l-1$, we keep track of all seams that were removed from $\F_{l}(\I)$. Denote $SC_l = \{ s_1, \dots{}, s_n\}$ as the set of all chosen seams at level $l$ (an example of one chosen seam is indicated in yellow in Figure~\ref{fig:deep_seam_carving_explain_a}).

\begin{figure}
\centering
\begin{subfigure}{0.35\textwidth}
  \centering
 \includegraphics[height=2.8cm]{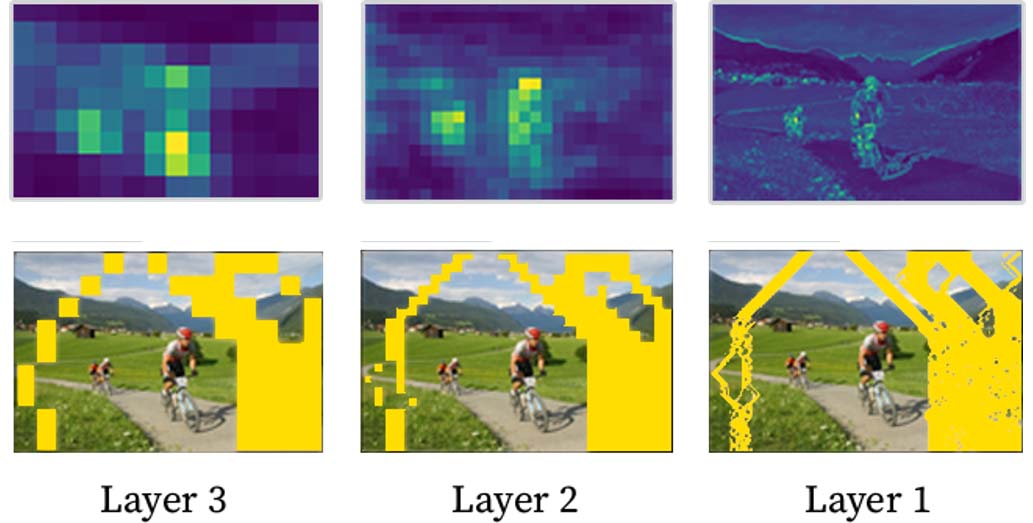}
  \caption{DSC.}
  \label{fig:deep_seam_bicycle_a}
\end{subfigure}
\begin{subfigure}{0.12\textwidth}
 \centering
 \includegraphics[height=2.8cm]{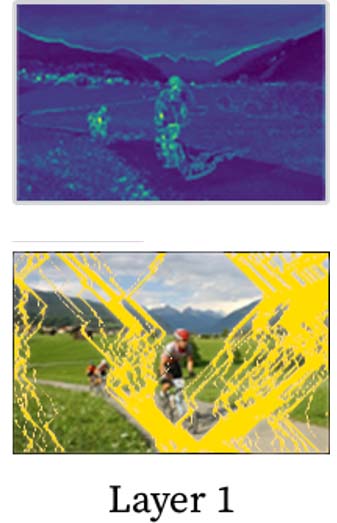}
  \caption{SC.}
  \label{fig:deep_seam_bicycle_b}
\end{subfigure}
\caption{Deep seam-carving vs. regular seam-carving. (a) Hierarchical deep seam-carving (DSC) applied on all three layers feature maps preserves the image's semantic information. (b) Results of original seam-carving (SC) using just the first layer feature-map as the importance map. Note how in this case, seams no longer avoid important regions.}
\label{fig:deep_seam_carving_bicycle}
\end{figure}

To find the minimal seams on $\F_{l-1}(\I)$, we consider a modified  importance map $MS_{l-1}$ at level $l-1$, that reduces the importance of regions that are part of the receptive field of the deep seams in level $l$. This attracts the seams at level $l-1$ to pass through the same regions and be consistent with the seams of level $l$ (see Figure~\ref{fig:deep_seam_carving_explain}).
The new map is given by the following equation: 
\begin{equation}
\label{eq:ModifiedL2Saliency}
    MS_{l-1}(i,j) = \begin{cases}
       \alpha \cdot S_{l-1}(i,j)& \quad \parbox{.135\textwidth}{$\exists s \in SC_{l}$ s.t $\F_{l-1}(i,j)$ is in the receptive field of $s$,}\\
       S_{l-1}(i,j) &\quad\text{otherwise} \\ 
     \end{cases}
\end{equation}
where $\alpha \in [0,1)$ is the scaling factor. Thus, the importance map in the finer layers inherit the information from the deeper layers, that implicitly constrain the selection of seams in the finer levels (Figure~\ref{fig:deep_seam_carving_bicycle}).

\paragraph{\textbf{Grid Warping}.}

Grid warping in image space is applied by first dividing the image to a grid of cells and then scaling each cell linearly using a different scaling factor. The scaling factors must adhere to the following two requirements. First, the total width of the scaled cells must match the target width $w'$. Second, the scaling factor of each cell should be proportional to the cell's importance. The first requirement guarantees that the resulting image size will match the target size, while the second requirement ensures less distortion in parts with high importance. 

For image width change, the initial width of each cell is given by $w_{G}$ and cells are assigned a \textit{scaling factor}, $\sigma_{i,j} \in [0,1]$, which specifies by how much each cell's width will be decreased. The actual resizing is applied using a linear scale - the width of cell $(i,j)$ is reduced by multiplying it by the scaling factor $\sigma_{i,j}$. In practice, it is useful to perform grid warping for width change by splitting the image into column-cells, defining only one cell in each column and one scaling factor $\sigma_i$. Otherwise, different cells in the same column may be distorted differently, which may lead to jittery results. 


To define the importance value $\mu_i$ of each column-cell $i$, we aggregate the importance-maps calculated by Equation~\ref{eq:L2SalMap} of all layers from the deepest until the first layer by up-sampling the deeper-layers maps to fit the size of the image. The values are then normalized to define the scaling factors as:
\begin{equation}
\label{eq:ScalingFactorEq}
\sigma_i = \frac{ w'}{ w_{G}} \cdot \frac{\mu_i}{\sum_{i} \mu_i}.
\end{equation}

\paragraph{\textbf{Deep Multi-Operator}.}

The combination of deep seam-carving and grid-warping is done by preventing deep seam-carving from removing seams with semantic content. To achieve this, we terminate the seam removal once the next seam's total importance is above a given threshold. However, we keep the same ratio of removed-seams to the original width of the feature-map in all layers, meaning that different number of seams are removed in each layer. Once deep seam-carving is terminated, and the image is reconstructed, we apply grid-warping on the intermediate resulting image to reach the final output in the desired size.

\subsection{Image Reconstruction}
\label{sec:Reonstruction}
Previous works show how to use a pre-trained CNN to synthesize images using back-propagation, for example to create images with different styles~\cite{Gatys16}. We adopt this approach, and use optimization to map back the target feature-maps into image-space, allowing us to obtain the resized output image. We use the target feature-maps to reconstruct our desired output by iteratively applying back-propagation to change the values of an image. Note that what we call output-image is in fact the input image to the network.

Our initial output image $\mathcal{O}$ is set to be a uniform 1D linear scaled version of the input image $\mathcal{I}$ (we consider other initialization methods, including random noise and seam carved image, and we compare the performance and results of each method in the supplementary materials). This allows the optimization to fix the distortions created by linear scale, and to re-construct the desired output by iteratively reducing distortions especially in important regions of the image (see Figure~\ref{fig:teaser}). Thus, we seek to update $\mathcal{O}$ by minimizing the total loss that is introduced by simple linear scale:
\begin{equation}
\label{eq:ReconstructionLoss}
\begin{split}
\mathcal{L} & =  \sum_{i=1}^{L} \lambda_i \cdot \norm{\F_{i}(\mathcal{O})  - \F'_{i}}_{2}, 
\end{split}   
\end{equation}
where $\F'_i$ are the $i^{th}$ layer target feature maps, and $\F_{i}(\mathcal{O})$ are the $i^{th}$ layer feature maps when the output image is fed into the pre-trained CNN. Moreover, $\lambda_1,\dots{},\lambda_L$ are non-negative hyper-parameters set to provide the weight of contribution of each term to the total loss. 

As suggested in~\cite{Gatys16}, minimizing the loss in Equation~\ref{eq:ReconstructionLoss} using gradient descent can produce visually pleasing images. In DNR, we use Adam Optimizer~\cite{Kingma14} to solve Equation~\ref{eq:ReconstructionLoss}.

\subsection{Image Refinement}
\label{sec:refinement}

The reconstruction optimization using back-propagation changes the pixel values of the output image $\mathcal{O}$ to minimize the loss function of 
Equation~\ref{eq:ReconstructionLoss}. This means that regions defined by the target feature-maps will most likely be preserved and reconstructed properly.
However, some artifacts such as checkerboard patterns and noisy pixels still appear in the resulting reconstructed image. These artifacts appear because 
content removed from the original image is causing discontinuities between better-preserved important regions and such locations accumulate gradients more than others (similar to artifacts created by de-convolution~\cite{odena2016deconvolution}).

\begin{figure}
    \centering
    \centerline{\includegraphics[width=8.0cm]{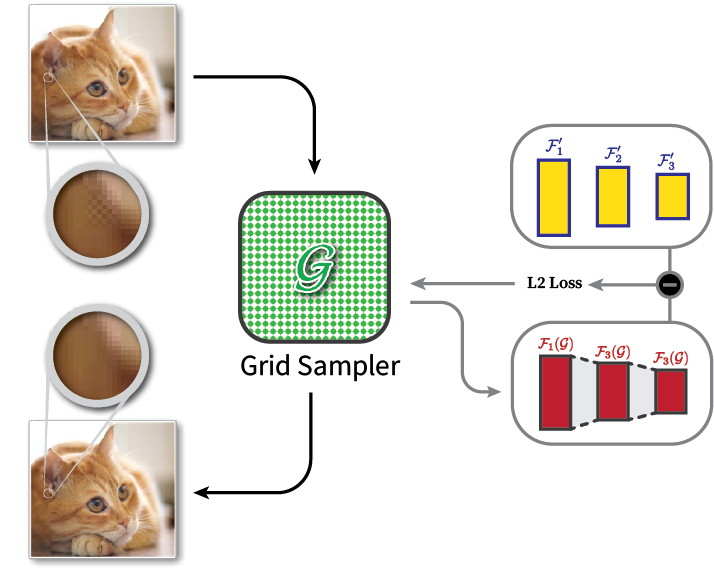}}
    \caption{The refinement procedure. The constructed image (left-top) after initial reconstruction (Section~\ref{sec:Reonstruction}) is fed into a grid sampler optimizing the same L2 loss function. The final output of DNR is the sampled reconstructed image (left-bottom).  }
    \label{fig:refinment_illus}
\end{figure}

We developed a novel method that utilized a grid-sampler layer $\mathcal{G}$ from~\cite{jaderberg2015spatial} to overcome these artifacts. Grid-sampling layer learns a mapping from positions of neurons in its input to positions in the output. In our case, we place such a layer as the first layer of the network, modifying the input to the network to be $\mathcal{G(O)}$ instead of $\mathcal{O}$ (see Figure~\ref{fig:refinment_illus}).

\begin{figure}
    \centering
    \includegraphics[width=8cm]{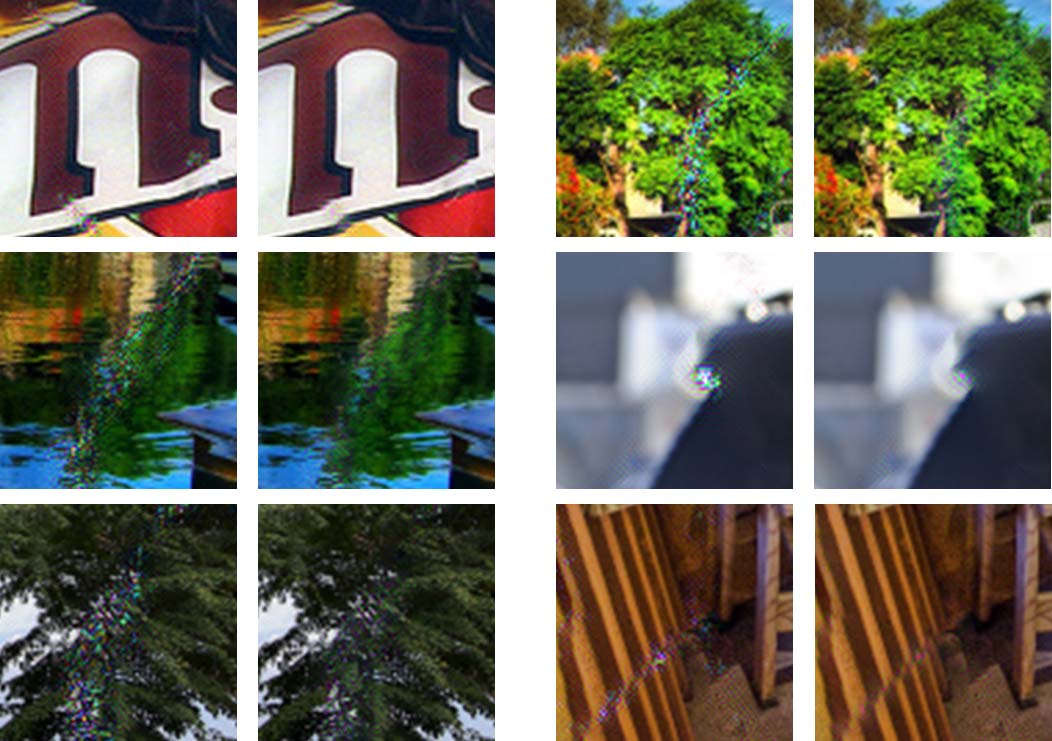}
    \caption{Refinement. Six large patches before (left) and after (right) the refinement procedure. Note the checkerboard and other artifacts before and after the refinement. 
    (zoom-in as needed).}
    \label{fig:stn_res}
\end{figure}

We use $\mathcal{G}$ only after the initial reconstruction of $\mathcal{O}$ is finalized (Section~\ref{sec:Reonstruction}). We add the grid-sampler layer and continue to optimize by using the same loss function of Equation~\ref{eq:ReconstructionLoss}. However, instead of changing the pixel values in $\mathcal{O}$, we keep them fixed and optimize the values of $\mathcal{G}$, the grid-sampler layer itself. In essence, this allows local shifts and interpolation of the pixels in  $\mathcal{O}$, causing the optimization to push and interpolate artifacts to near-by edges and overcome unpleasant checkerboard artifacts (see Figure~\ref{fig:stn_res}).
The final output of DNR, i.e. the resized image, is the application of the grid-sampler on the reconstructed image, i.e.\ $\mathcal{G(O)}$.

\section{Results}

In our experiments, we use VGG19~\cite{Simonyan14}, which was trained on ImageNet~\cite{ILSVRC15} dataset. Throughout this section, we use selected ReLU activation and Max-Pooling activation  in VGG19's layers as our feature maps $\F_i(\I)$
Denote $block_i\_conv_j$ as the ReLU activation of the $j$-th convolution layer in block $i$, and $block_i\_pool$ as the pooling activation of block $i$.  
The default configuration of our experimental results, unless otherwise stated, uses $block_1\_conv_2$, $block_2\_conv_2$, $block_3\_conv_4$, $block_4\_conv_5$ and $block_5\_pool$ as feature maps. 

We always remove at least one seam in the deepest feature map, and remove more seams only if their importance are within the 20-$th$ percentile of the importance map. The value of the parameters used in the reconstruction loss (Equation~\ref{eq:ReconstructionLoss}) are $\lambda_{1} = 1$ and for $i>1$, $\lambda_{i} = 0$. The scaling factor in Equation~\ref{eq:ModifiedL2Saliency} is set to $\alpha=0.5$. Finally, the grid size we use for warpping is $16$.

\subsection{Importance Map Effectiveness}
\label{EffectiveMap}

\begin{figure}
\centering
    \begin{subfigure}{0.45\textwidth}
    \centering
    \adjincludegraphics[width=0.305\textwidth]{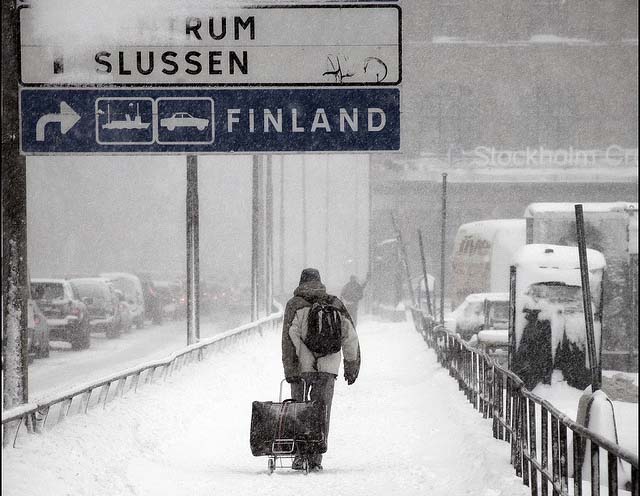}
    \adjincludegraphics[width=0.305\textwidth]{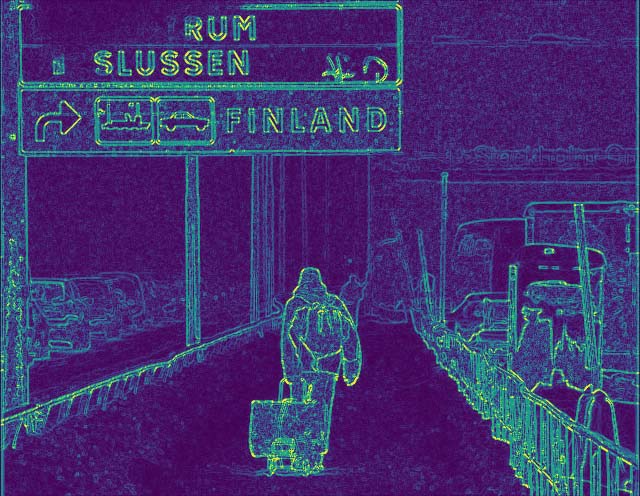}
    \adjincludegraphics[width=0.305\textwidth]{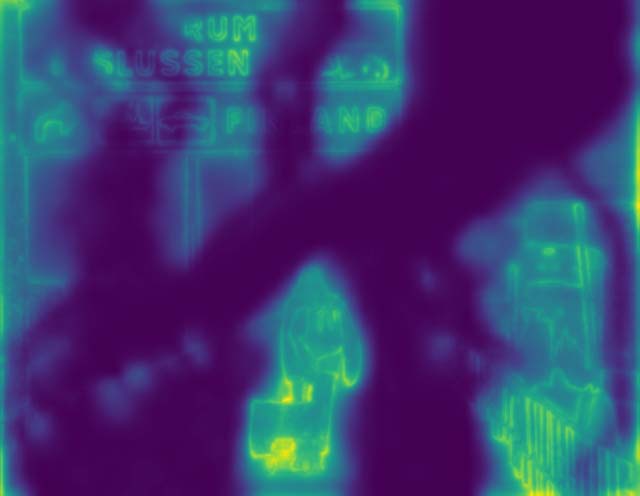}
    \end{subfigure}
    
    \begin{subfigure}{0.45\textwidth}
    \centering
    \adjincludegraphics[width=0.305\textwidth]{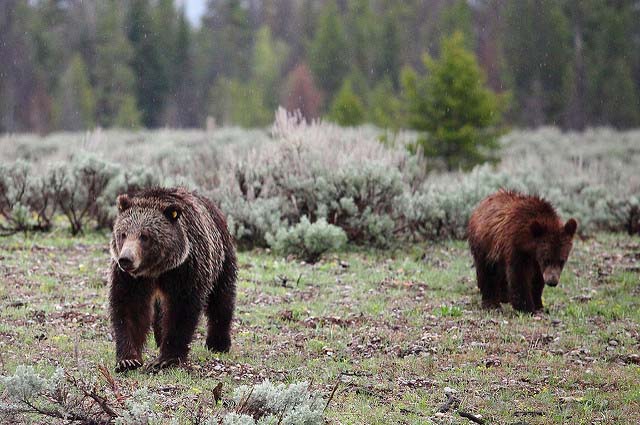}
    \adjincludegraphics[width=0.305\textwidth]{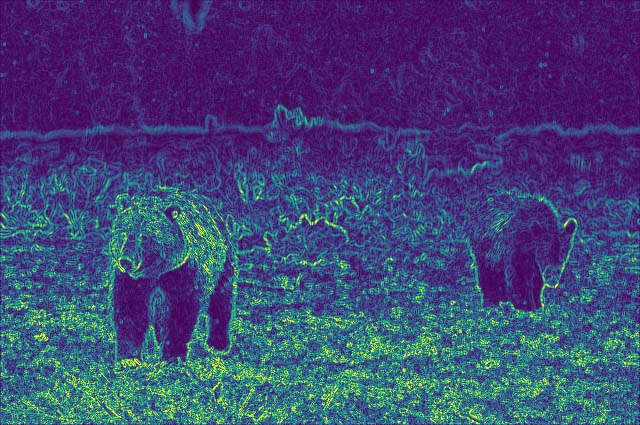}
    \adjincludegraphics[width=0.305\textwidth]{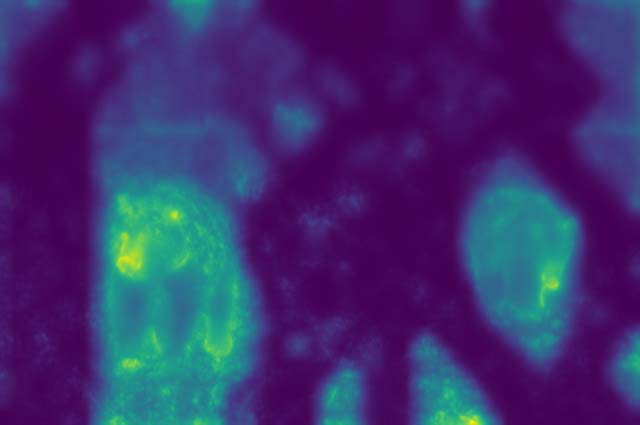}
    \end{subfigure}
    
    
    
    
    \begin{subfigure}{0.45\textwidth}
    \centering
    \adjincludegraphics[width=0.305\textwidth]{images/effective_importance_map/b_1.jpg}
    \adjincludegraphics[width=0.305\textwidth]{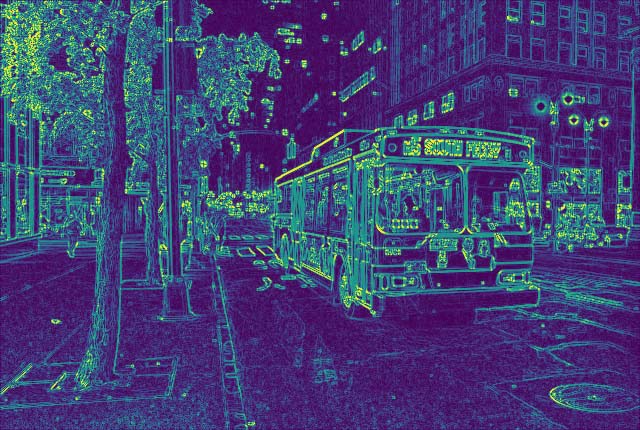}
    \adjincludegraphics[width=0.305\textwidth]{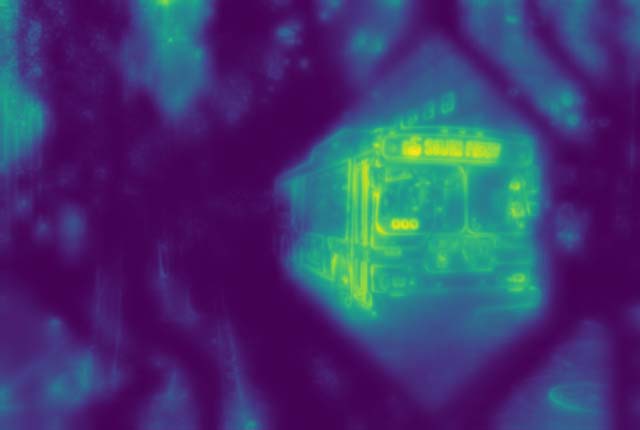}
    \end{subfigure}

    \begin{subfigure}{0.45\textwidth}
    \vspace{0.01\textwidth}
	\hspace{0.15 \textwidth} \text{(a)}
	\hspace{0.25\textwidth} \text{(b)}
	\hspace{0.25\textwidth} \text{(c)}
    \end{subfigure}
	\caption{Visualization of the importance map as seen by Deep Seam Carving. We show the input importance map (b) that is used in Seam Carving \cite{Avidan07}, and the effective importance map used by Deep Seam Carving (c). The effective importance map is more focused on semantic areas, which suggests less distortion to important regions.}
	\label{fig:effective_importance_map_comparison}
\end{figure}

The importance map used in the original Seam Carving algorithm~\cite{Avidan07} is based on gradient magnitude of the image. This map is often used as the base importance for many other retargeting algorithms as well.
In Figure~\ref{fig:effective_importance_map_comparison}, we compare this importance map, and the effective importance map we use for Deep Seam Carving.  The effective importance map is derived by summing of importance maps used by Deep Seam Carving. To visualize the importance map, we up-sample low-resolution maps to match the image shape. As can be seen, the original importance map tends to concentrate on edges and lacks the ability to capture semantics, while our map clearly gives higher importance to semantic objects in the image.

\subsection{Feature Space vs Image Space}\label{SeamCarvingImageVsFeature}

\begin{figure}
\centering
\begin{subfigure}{0.225\textwidth}
  \centering
  \includegraphics[scale=0.5]{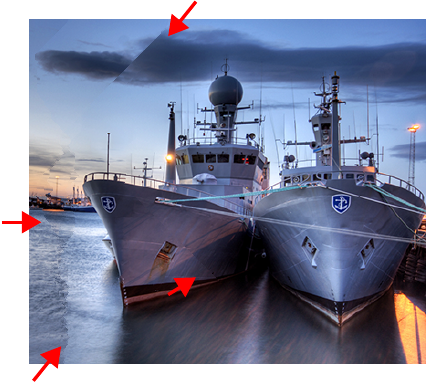}
  \caption{Image space.}
  \label{fig:carving_image_vs_feature_a}
\end{subfigure}
\begin{subfigure}{0.222\textwidth}
  \centering
  \includegraphics[scale=0.5]{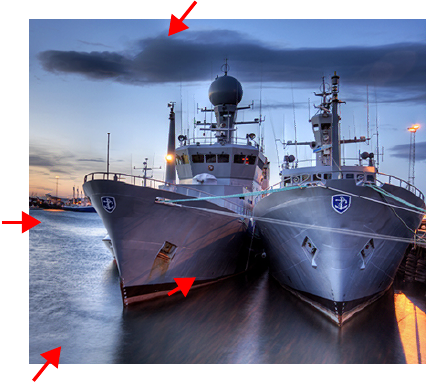}
  \caption{Feature space.}
  \label{fig:carving_image_vs_feature_b}
\end{subfigure}
\caption{ Removing seams from the input image results in discontinuous regions (a), while reconstruction using DNR produces better results (b).}
\label{fig:carving_image_vs_feature}
\end{figure}

A possible alternative approach that will still use deep feature maps would be to apply seam carving in image space while using the feature maps as importance maps. Therefore, instead of removing seams from the feature maps, one can consider removing the same seams from the input image in order to produce the output image. 

Figure~\ref{fig:carving_image_vs_feature} shows a comparison of this approach to DNR that is based on reconstruction. As can be seen, image space retargeting leads to artifacts due to removing many seams from the same region. In contrast, in our DNR method, reconstructing the image leads to more continuous results. First because neighboring activations in VGG19 have overlapping receptive fields, thus affecting several output pixels in the reconstruction. Second, because using a CNN that that was trained on natural images tends to generate photo-realistic images as well.

\subsection{Reconstruction via Deep Layers Activation}
\label{ReconstructionViaDeepLayers}
In the following experiments we compare between different initialization methods and consider different values for the hyper parameters $\{\lambda_i \}$. The experiments were conducted without the refinement procedure, so we can better compare the quality of the reconstructed images. Moreover, we extract $block_{1}\_conv_{1}$, $block_{3}\_conv_{1}$, and $block_{5}\_conv_{1}$ activation and use them as our feature maps.

\begin{figure*}
    \centering
    \begin{subfigure}{0.33\textwidth}
    \centering
     \includegraphics[width=5cm]{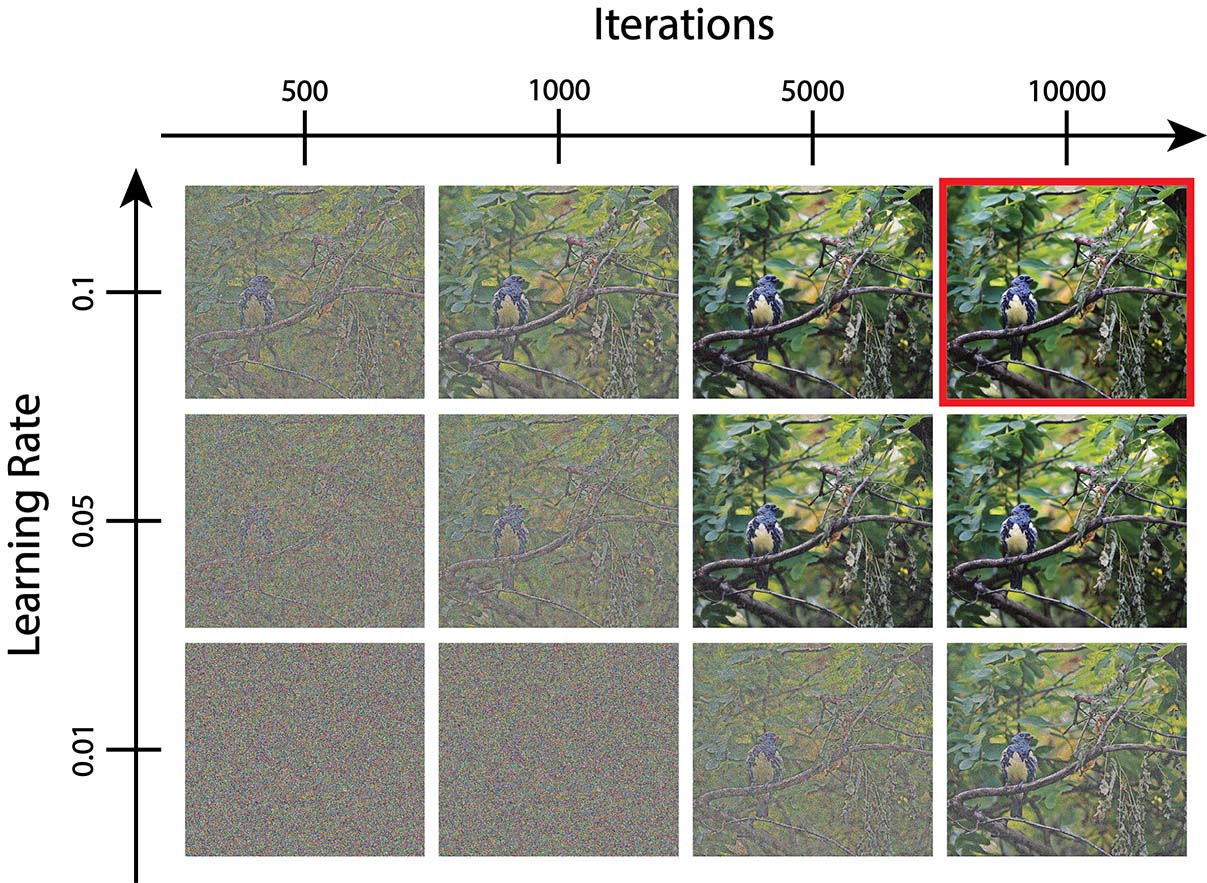}
     \caption{Random Initialization}
    \end{subfigure}
    \begin{subfigure}{0.3\textwidth}
    \centering
     \includegraphics[width=5cm]{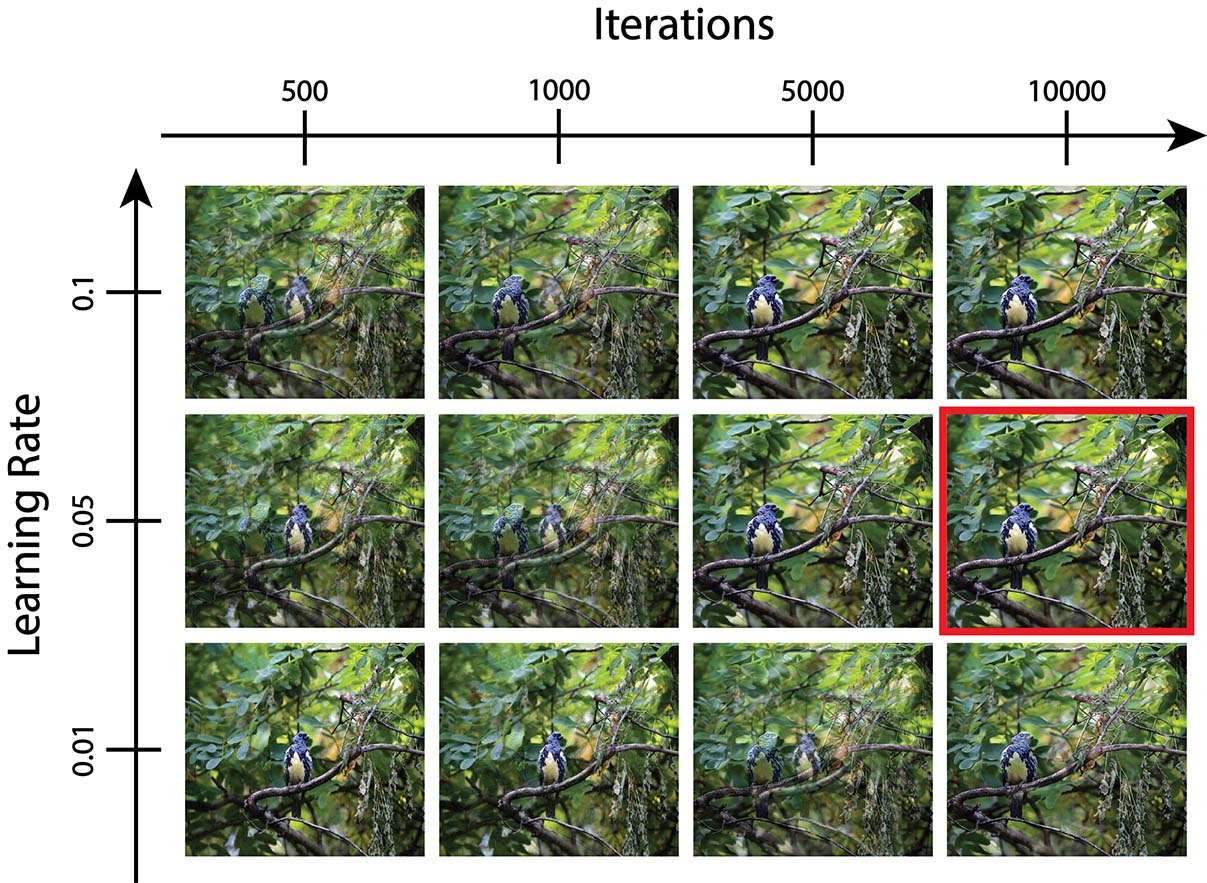}
    \caption{Linear Scale Initialization}
    \end{subfigure}
    \begin{subfigure}{0.33\textwidth}
    \centering
     \includegraphics[width=5cm]{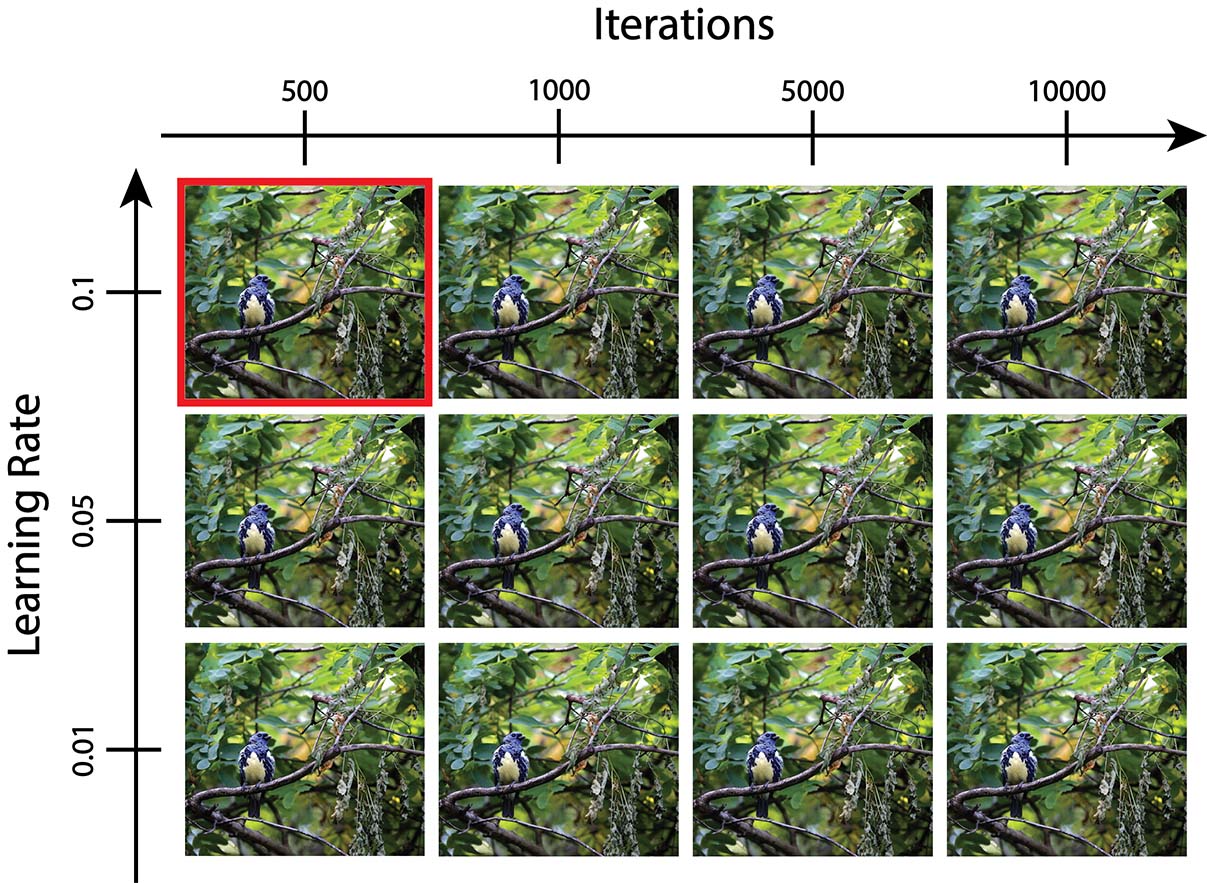}
    \caption{Seam Carving Initialization}
    \end{subfigure}
    \caption{Different initialization methods. We compare different initialization methods for the output image during the reconstruction phase, including (a) random noise, (b) linear scale and (c) seam carving. In each row we show the output for a given learning rate, and in each column we show a snapshot of the output at a certain iteration. The highlighted output is the best outcome for each initialization method (zoom in if needed).}
    \label{fig:init_method}
\end{figure*}

In Figure~\ref{fig:init_method}, we compare different initialization methods for the initial solution of the reconstruction phase. Specifically, we consider three methods; random initialization, linear scale and seam carving. In random initialization the solution's pixels values are uniformly sampled, while in linear scale initialization the input image is scaled to the desired shape using 1D uniform scale. Finally, in seam carving initialization, we remove the same seams that were removed from low-level feature maps (i.e $block_{1}\_conv_{1}$).

As can be seen, with random initialization the output contains many artifacts. Nonetheless, we are able to produce visually pleasing images for both seam carving and linear scale initialization. For linear scale it is best to use low learning rates while increasing the maximum number of iterations. For seam-carving initialization, we need far fewer optimization steps in order to achieve high-quality images.

\begin{figure*}
    \centering
    \begin{subfigure}{0.35\textwidth}
    \centering
     \includegraphics[height=4.cm]{images/deep_reconstruction/umbrella2.jpg}
     \caption{Input Image}
    \end{subfigure}
    \begin{subfigure}{0.27\textwidth}
    \centering
     \includegraphics[height=4.cm]{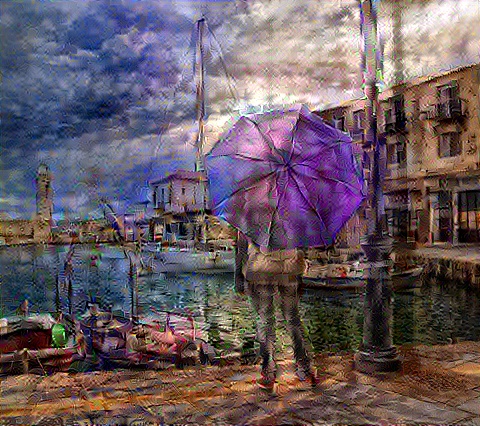}
     \caption{Linear Scale Initialization}
    \end{subfigure}
    \begin{subfigure}{0.27\textwidth}
    \centering
     \includegraphics[height=4.cm]{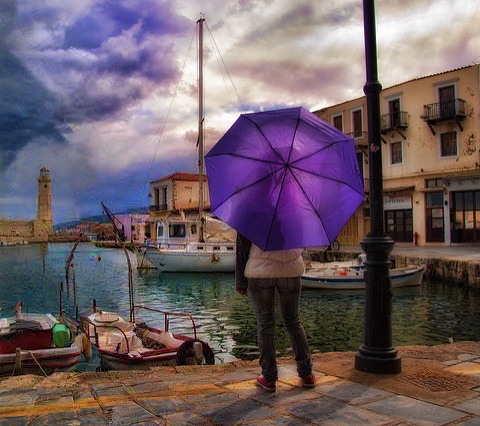}
     \caption{Seam Carving Initialization}
    \end{subfigure}
    \caption{Reconstruction using deep layers only ($\lambda_1 = \lambda_2 = 0.0$ and $\lambda_3=1.0$). We scale the input image (a) width by 75\% using Deep Seam carving and compare the reconstructed image for different initialization methods (b)-(c). As can be seen, seam carving initialization (c) produces higher-quality images while highly distorted image is produced when linear scale initialization (b) is used.}
    \label{fig:deep_reconst_fig}
\end{figure*}
Next, we show the contribution of deep layers to the quality of the output image, and consider various values of $\{ \lambda_1, \lambda_2, \lambda_3\}$. In Figure~\ref{fig:deep_reconst_fig} we show the output image when $\{ \lambda_i\} = \{0.0,0.0,1.0\}$. As shown, in linear scale initialization the output is highly-distorted and the optimization doesn't converge to a visually-pleasing solution. However, if we use seam carving initialization we get better results. 

These findings are expected for two reasons. First, synthesizing photo-realistic images from deep layers activation is hard (as suggested in the paper of Gatys \etal~\cite{Gatys16}). Second, in seam carving initialization, our solution is already photo-realistic except at regions where seams were removed. Thus, few pixels need to be altered in order to reduce the loss function. On the other hand, in linear scale initialization the initial loss is relatively-high, and many pixels are changed between optimization steps. 

\begin{figure*}
    \centering
    \begin{subfigure}{0.27\textwidth}
    \centering
     \includegraphics[height=4cm]{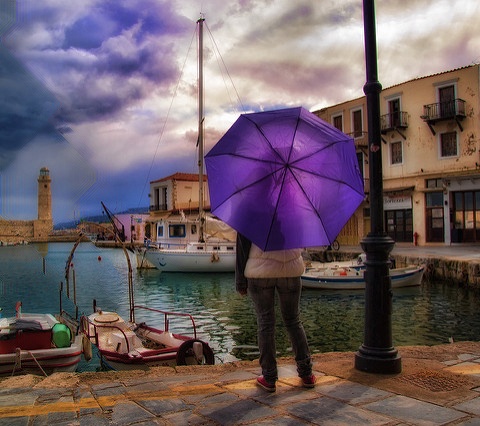}
     \caption{$\lambda_1 = 1.0$ and $\lambda_2 =\lambda_3= 0.0$}
    \end{subfigure}
    \begin{subfigure}{0.27\textwidth}
    \centering
     \includegraphics[height=4cm]{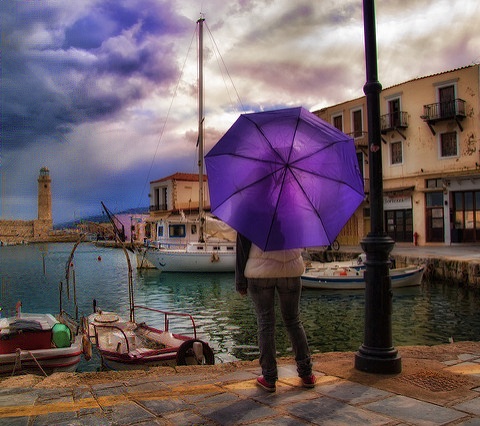}
     \caption{$\lambda_1 = \lambda_2 = 0.5$ and $\lambda_3= 0.0$}
    \end{subfigure}
    \begin{subfigure}{0.27\textwidth}
    \centering
     \includegraphics[height=4cm]{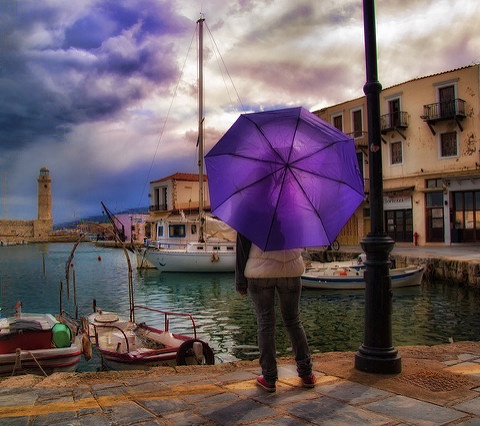}
     \caption{$\lambda_1 = \lambda_2 = \lambda_3 = 0.33$}
    \end{subfigure}
    \begin{subfigure}{0.07\textwidth}
    \centering
     \includegraphics[height=4cm]{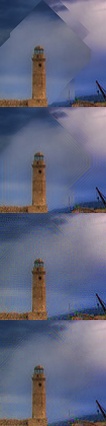}
     \caption{}
    \end{subfigure}
    \caption{Reconstruction with seam carving initialization for different $\lambda$. In (a)-(c) we see improvement when considering deep layers in the reconstruction loss. A comparison between image patches (d) is shown as well, where the top patch is the initial solution without reconstruction, and the three patches are cropped from the images in (a)-(c) respectively (zoom-in if needed).}
    \label{fig:sc_init_reconst}
\end{figure*}
\begin{figure*}
    \centering
    \begin{subfigure}{0.27\textwidth}
    \centering
     \includegraphics[height=4cm]{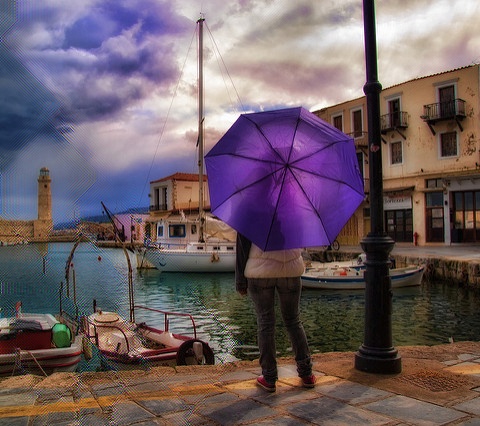}
     \caption{$\lambda_1 = 1.0$ and $\lambda_2 =\lambda_3= 0.0$}
    \end{subfigure}
    \begin{subfigure}{0.27\textwidth}
    \centering
     \includegraphics[height=4cm]{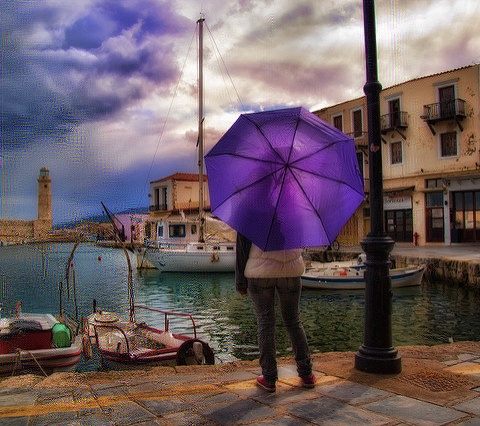}
     \caption{$\lambda_1 = \lambda_2 = 0.5$ and $\lambda_3= 0.0$}
    \end{subfigure}
    \begin{subfigure}{0.27\textwidth}
    \centering
     \includegraphics[height=4cm]{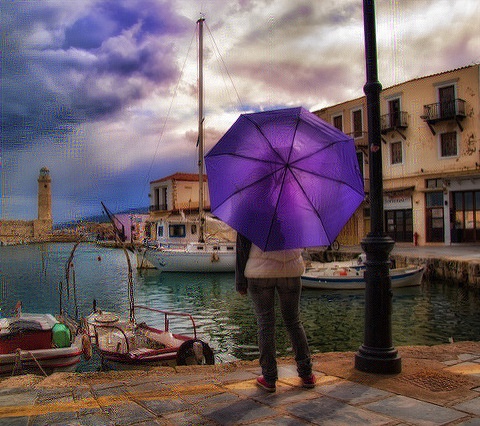}
     \caption{$\lambda_1 = \lambda_2 = \lambda_3 = 0.33$}
    \end{subfigure}
    \begin{subfigure}{0.07\textwidth}
    \centering
     \includegraphics[height=4cm]{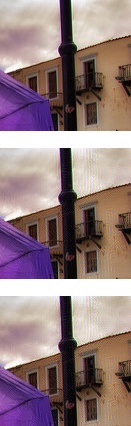}
     \caption{}
    \end{subfigure}
    \caption{Reconstruction with linear scale initialization for different $\lambda$. In (a)-(c) we see deterioration in the image quality when considering deep layers in the reconstruction loss. A comparison between image patches (d) is shown as well, where the top patch is from the image in (a) and the second patch is from (b) and the third and last patch  is from (c) (zoom-in if needed).}
    \label{fig:ls_init_reconst}
\end{figure*}

Finally, in Figure~\ref{fig:sc_init_reconst} and Figure~\ref{fig:ls_init_reconst} we show the contribution of deep layers to the quality output image. In Figure~\ref{fig:sc_init_reconst}, we see that deep layers contribute to the quality of the output image, while in linear scale initialization (Figure~\ref{fig:ls_init_reconst}, deep layers negatively contribute to the quality of the output image.

\subsection{Visual Comparison with Previous Methods}
\label{QualitativeEvaluation}
\begin{figure*}
\centering
    \begin{subfigure}{0.33\textwidth}
    \centering
    \adjincludegraphics[width=2.2cm, height=2.4cm]{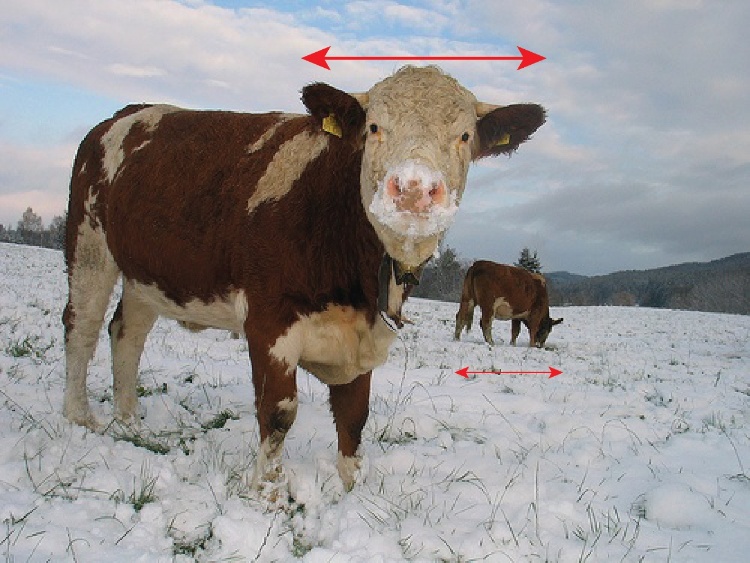}
    \hspace{0.1cm}
    \adjincludegraphics[width=1.1cm,height=2.4cm]{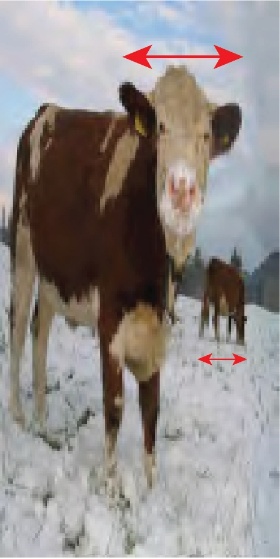}
    \hspace{0.1cm}
    \adjincludegraphics[width=1.1cm,height=2.4cm]{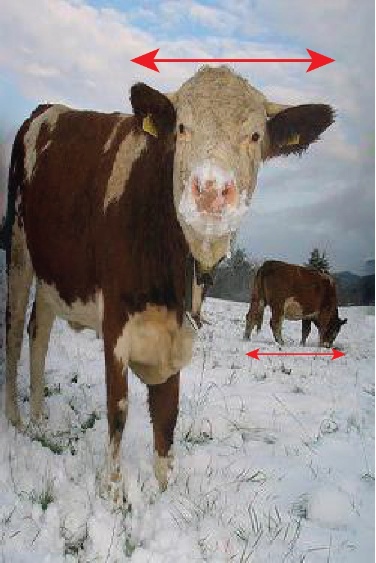}
    \end{subfigure}
    \begin{subfigure}{0.33\textwidth}
    \centering
    \adjincludegraphics[width=2.2cm, height=2.4cm]{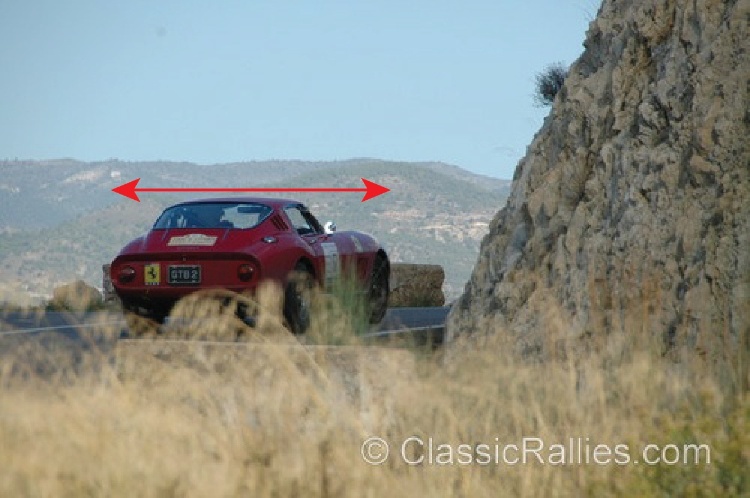}
    \hspace{0.1cm}
    \adjincludegraphics[width=1.1cm,height=2.4cm]{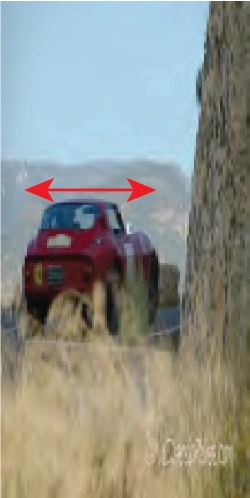}
    \hspace{0.1cm}
    \adjincludegraphics[width=1.1cm,height=2.4cm]{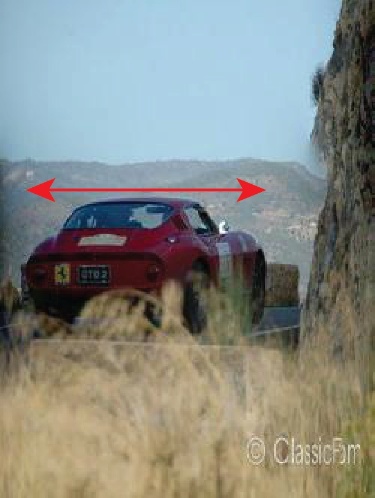}
    \end{subfigure}
    \begin{subfigure}{0.33\textwidth}
    \centering
    \adjincludegraphics[width=2.2cm, height=2.4cm]{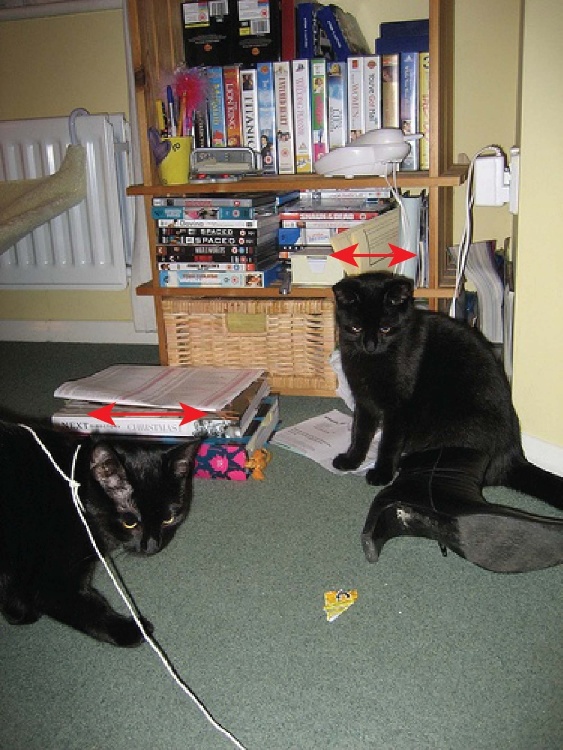}
    \hspace{0.1cm}
    \adjincludegraphics[width=1.1cm,height=2.4cm]{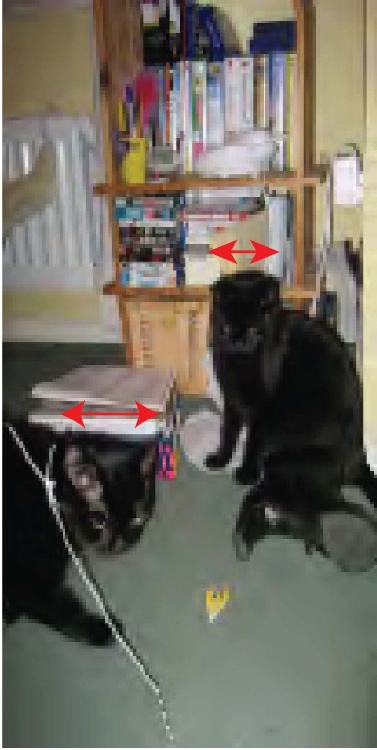}
    \hspace{0.1cm}
    \adjincludegraphics[width=1.1cm,height=2.4cm]{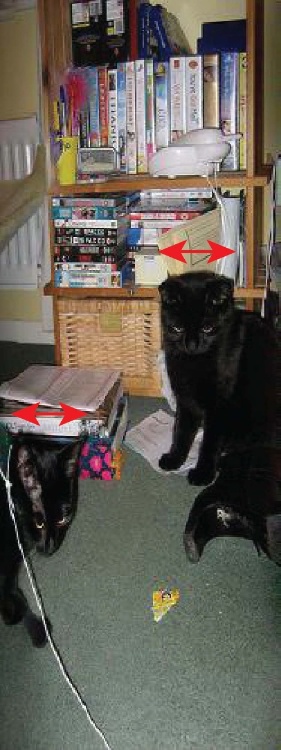}
    \end{subfigure}
	\caption{50\% width scale. The input images (left) are from The Pascal VOC2017 dataset~\cite{pascal-voc-2007}. We compare results of WSSDCNN~\cite{Cho17} (middle) and DNR (right). The results are obtained by setting $\alpha = 0.2$ and employing Deep Seam Carving to perform 50\% of the retargeting task. As can be seen, DNR better preserves the images subjects (see guidelines).}
	\label{fig:ours_vs_wssdcnn}
\end{figure*}
\begin{figure*}
\centering
    \begin{subfigure}{0.33\textwidth}
    \centering
    \adjincludegraphics[height=1.5cm]{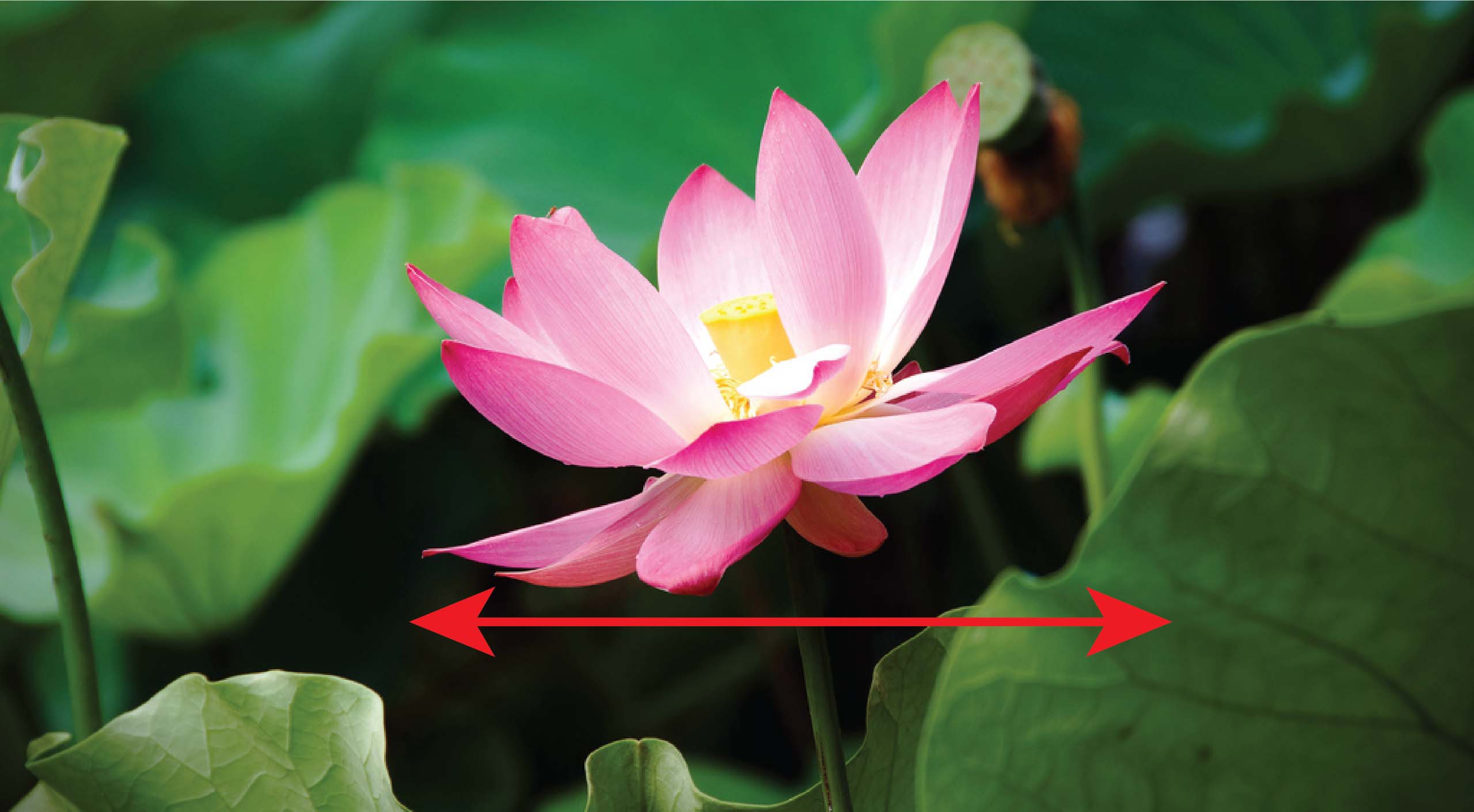}
    \hspace{0.05cm}
    \adjincludegraphics[height=1.5cm]{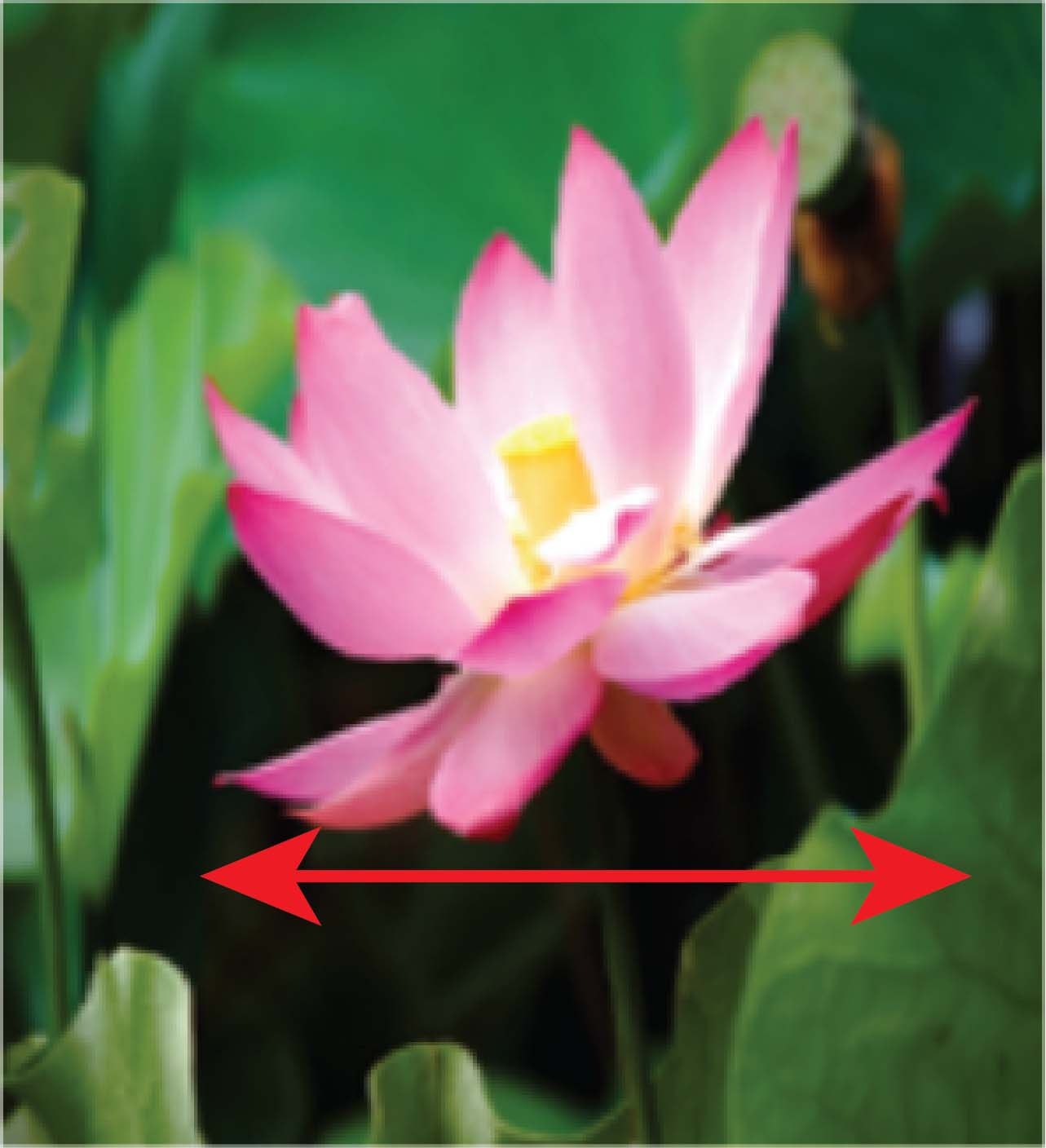}
    \adjincludegraphics[height=1.5cm]{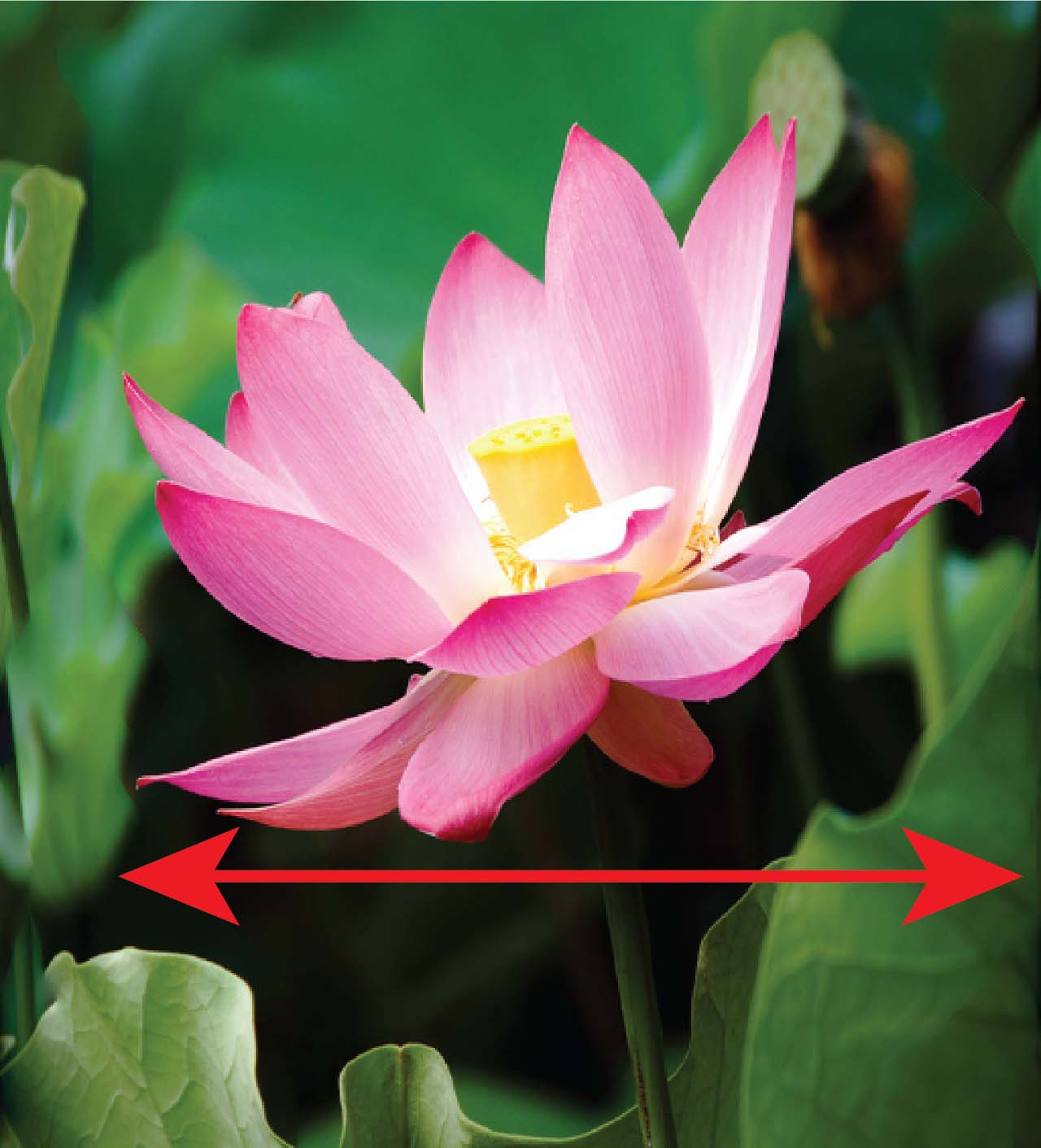}
    \end{subfigure}
    \begin{subfigure}{0.33\textwidth}
    \centering
    \adjincludegraphics[height=1.5cm]{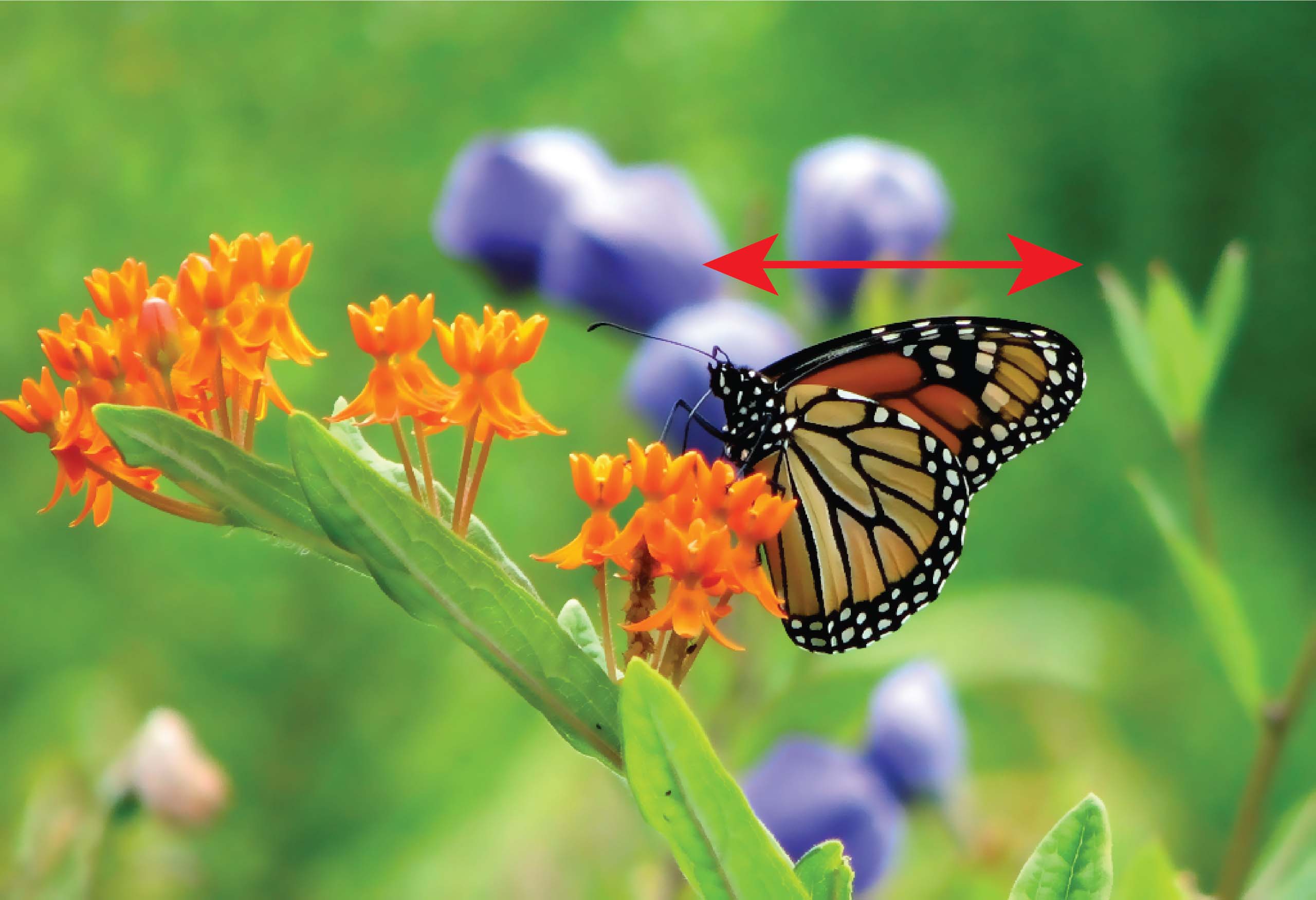}
    \hspace{0.05cm}
    \adjincludegraphics[height=1.5cm]{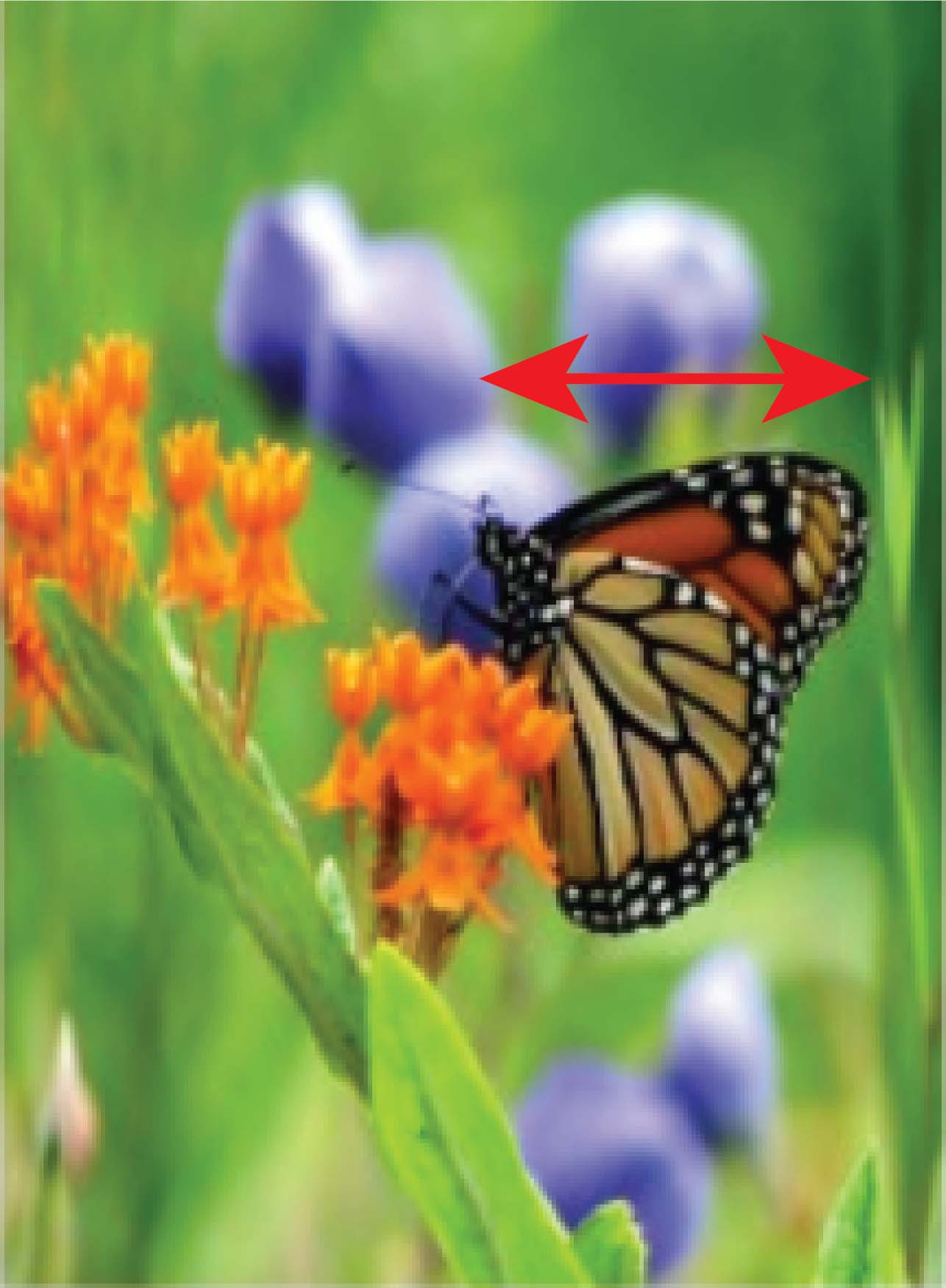}
    \hspace{0.05cm}
    \adjincludegraphics[height=1.5cm]{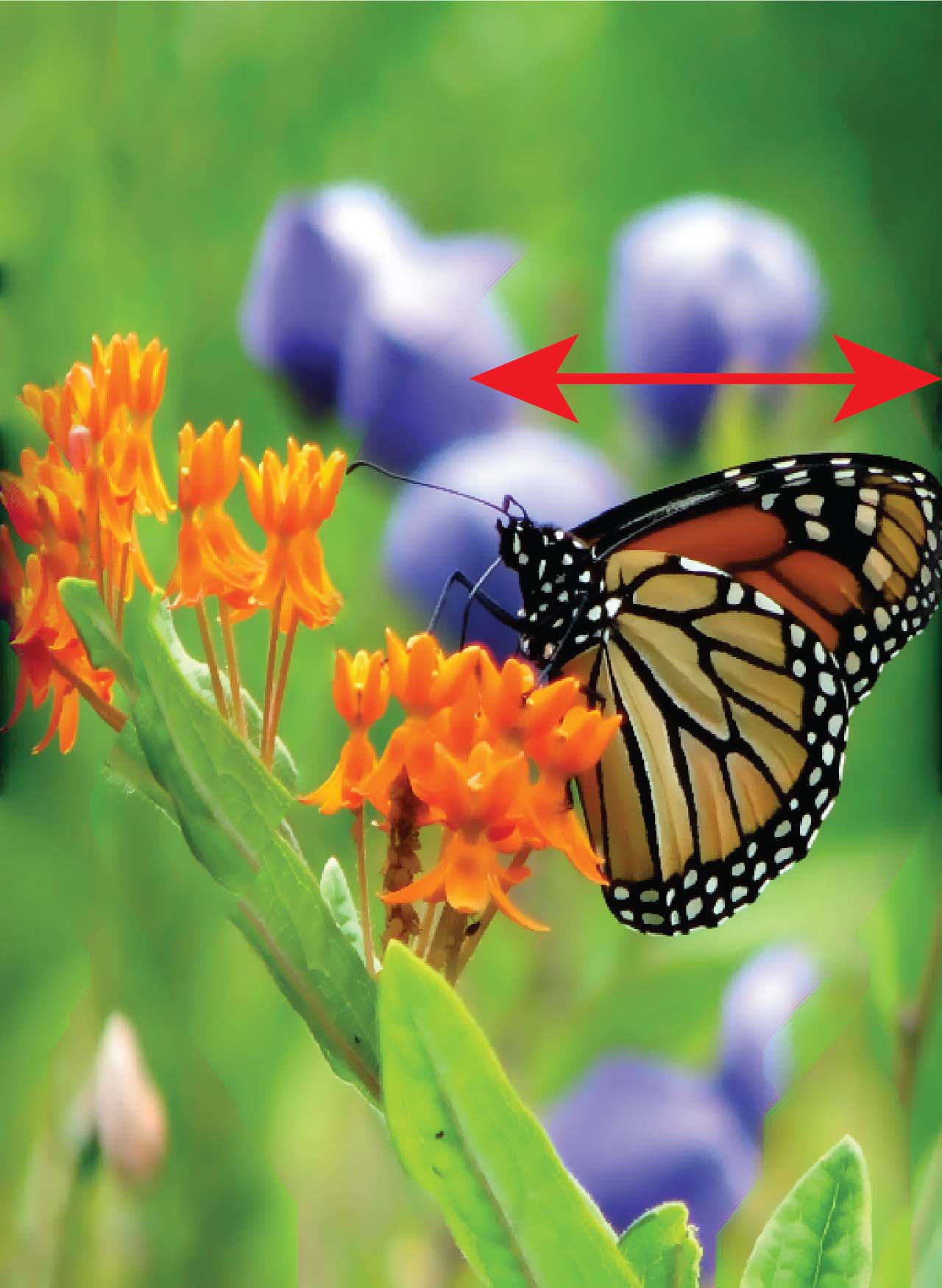}
    \end{subfigure}
    \begin{subfigure}{0.33\textwidth}
    \centering
    \adjincludegraphics[height=1.5cm]{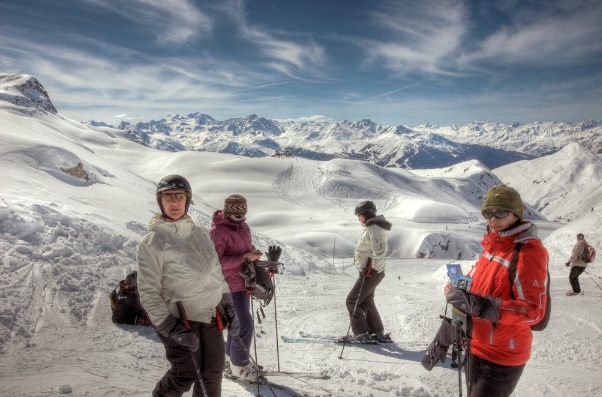}
    \hspace{0.05cm}
    \adjincludegraphics[height=1.5cm]{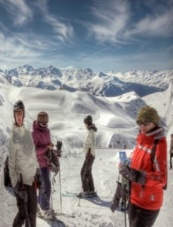}
    \hspace{0.05cm}
    \adjincludegraphics[height=1.5cm]{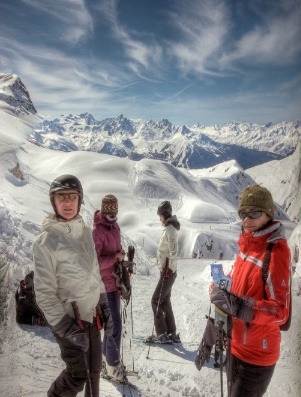}
    \end{subfigure}
    
	\caption{50\% width scale. (left) Input image from RetargetMe \cite{Rubinstein10Comparative}, (middle) results in \cite{DBLP:journals/corr/abs-1811-07793} and (right) DNR. We use $\alpha = 0.2$ and only remove seams if their importance are within the $35$-percentile of the importance map. As can be seen, DNR better preserves important regions. To see this, please notice the original width of the salient subjects and compare them with each of the retargeted results (see guidelines).}
	\label{fig:ours_vs_deepir}
\end{figure*}
\begin{figure*}
\centering
    \begin{subfigure}{\textwidth}
    \centering
	\adjincludegraphics[width=0.175\textwidth]{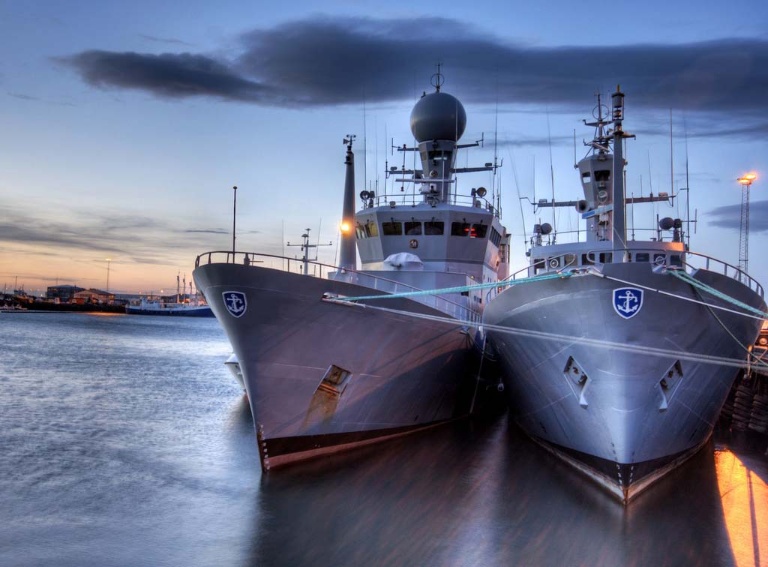}  
	\adjincludegraphics[width=0.13125\textwidth]{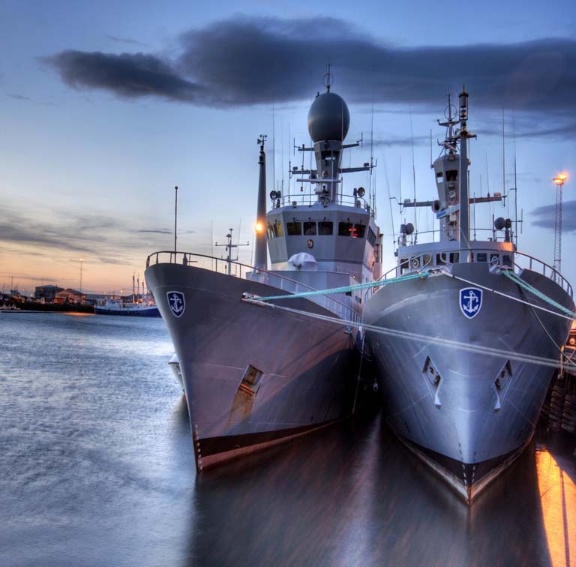}
    \adjincludegraphics[width=0.13125\textwidth]{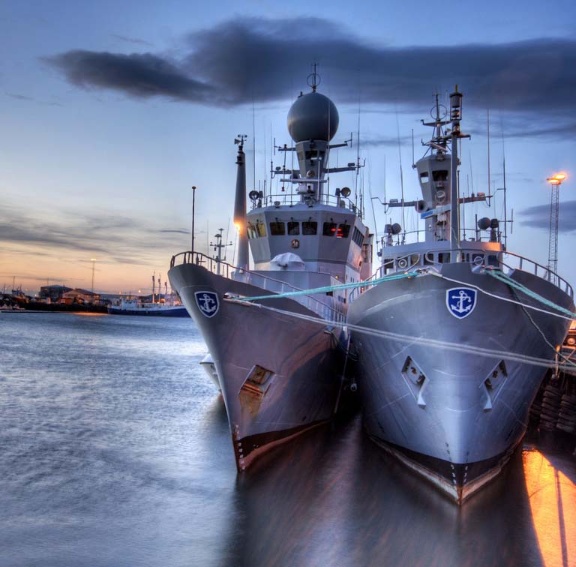}
    \adjincludegraphics[width=0.13125\textwidth]{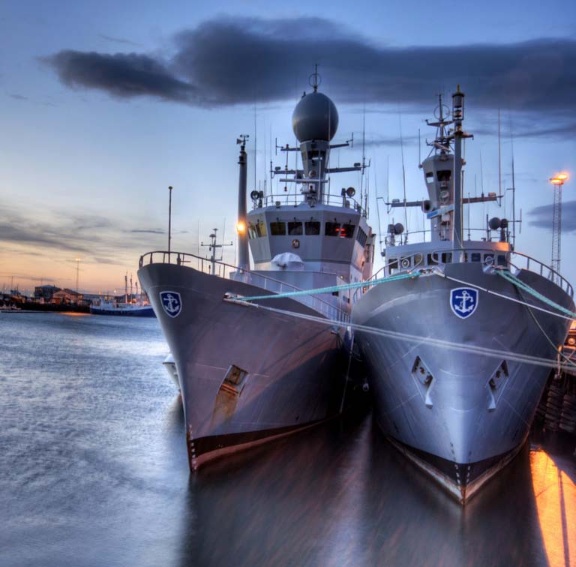}
    \adjincludegraphics[width=0.13125\textwidth]{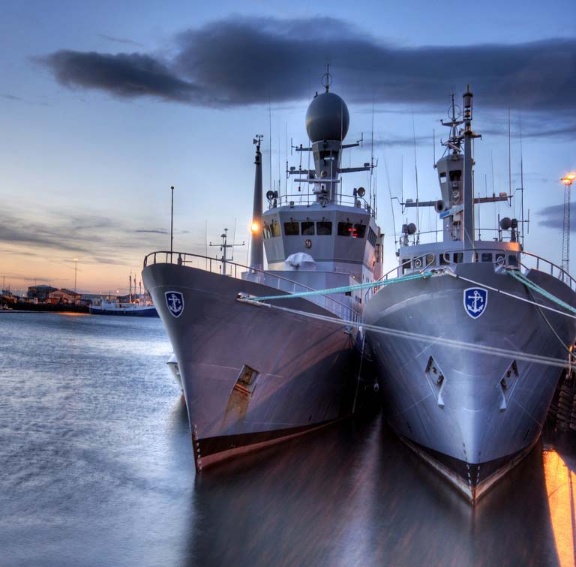}
    \adjincludegraphics[width=0.13125\textwidth]{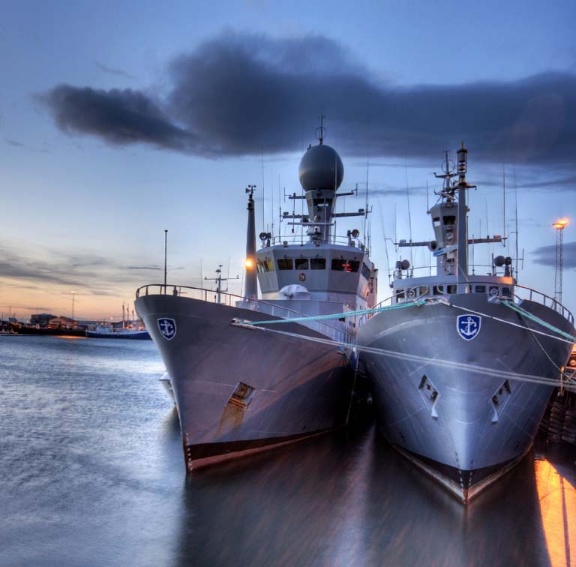}
    \adjincludegraphics[width=0.13125\textwidth]{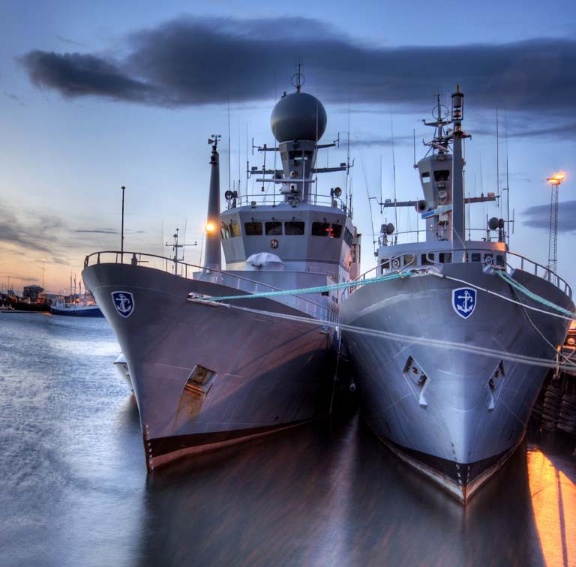}
    \end{subfigure}
    \begin{subfigure}{\textwidth}
    \centering
	\adjincludegraphics[width=0.175\textwidth]{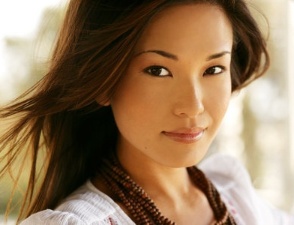}  
	\adjincludegraphics[width=0.13125\textwidth]{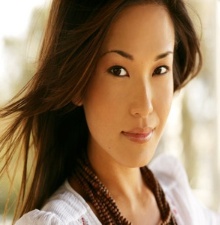}
    \adjincludegraphics[width=0.13125\textwidth]{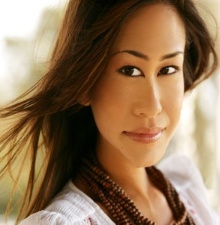}
    \adjincludegraphics[width=0.13125\textwidth]{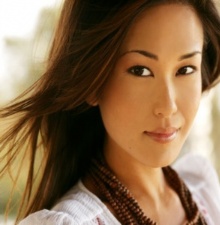}
    \adjincludegraphics[width=0.13125\textwidth]{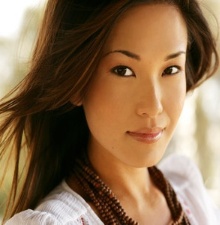}
    \adjincludegraphics[width=0.13125\textwidth]{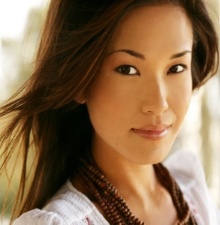}
    \adjincludegraphics[width=0.13125\textwidth]{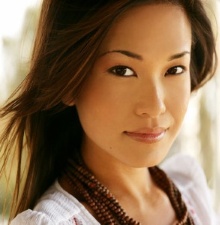}
    \end{subfigure}
 
    \begin{subfigure}{\textwidth}
    \vspace{0.01\textwidth}
	\hspace{0.08\textwidth} \text{(a)}
	\hspace{0.12\textwidth} \text{(b)}
	\hspace{0.105\textwidth} \text{(c)}
	\hspace{0.105\textwidth} \text{(d)}
	\hspace{0.105\textwidth} \text{(e)}
	\hspace{0.105\textwidth} \text{(f)}
	\hspace{0.105\textwidth} \text{(g)}
    \end{subfigure}

	\caption{Scaling the width of the input image by 75\%. (a) Input image, (b) Linear Scale, (c) Seam Carving~\cite{Rubinstein08}, (d) Warping~\cite{Wolf07}, (e) Multiop~\cite{Rubinstein09}, (f) SV~\cite{Krahenbuhl09} and (g) DNR (ours).}
	\label{fig:retarget_me_comparison_75}
\end{figure*}
\begin{figure*}
\centering
    \begin{subfigure}{\textwidth}
    \centering
    \adjincludegraphics[width=0.24\textwidth]{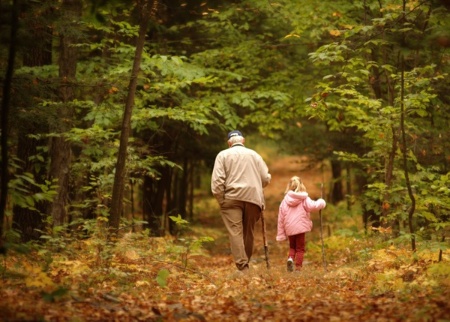}
    \adjincludegraphics[width=0.12\textwidth]{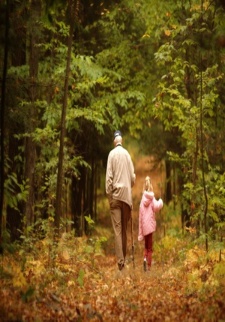}
    \adjincludegraphics[width=0.12\textwidth]{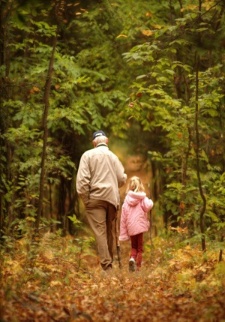}
    \adjincludegraphics[width=0.12\textwidth]{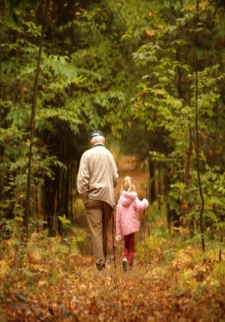}
    \adjincludegraphics[width=0.12\textwidth]{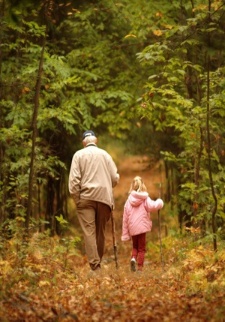}
    \adjincludegraphics[width=0.12\textwidth]{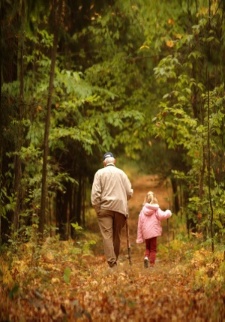}
    \adjincludegraphics[width=0.12\textwidth]{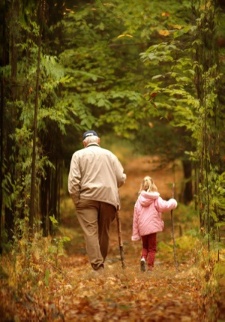}
    \end{subfigure}
    \begin{subfigure}{\textwidth}
    \centering
    \adjincludegraphics[width=0.24\textwidth]{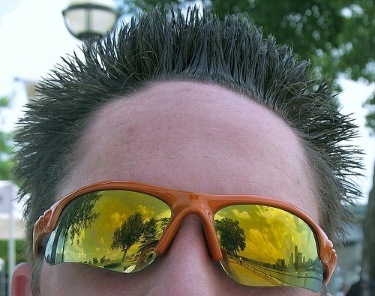}
    \adjincludegraphics[width=0.12\textwidth]{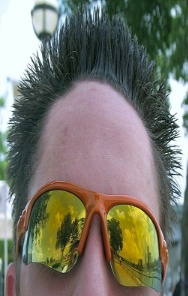}
    \adjincludegraphics[width=0.12\textwidth]{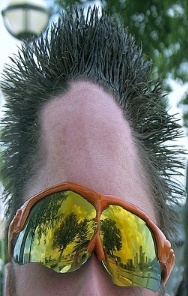}
    \adjincludegraphics[width=0.12\textwidth]{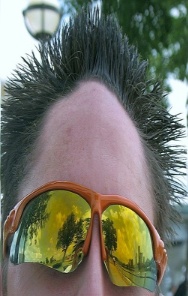}
    \adjincludegraphics[width=0.12\textwidth]{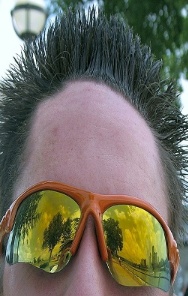}
    \adjincludegraphics[width=0.12\textwidth]{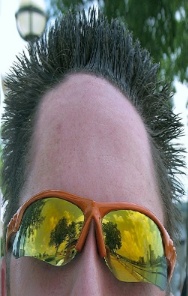}
    \adjincludegraphics[width=0.12\textwidth]{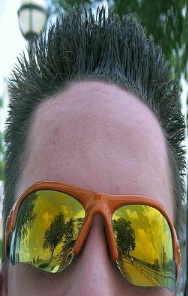}
    \end{subfigure}

    \begin{subfigure}{\textwidth}
    \vspace{0.01\textwidth}
	\hspace{0.12\textwidth} \text{(a)}
	\hspace{0.15\textwidth} \text{(b)}
	\hspace{0.0925\textwidth} \text{(c)}
	\hspace{0.0925\textwidth} \text{(d)}
	\hspace{0.0925\textwidth} \text{(e)}
	\hspace{0.0925\textwidth} \text{(f)}
	\hspace{0.0925\textwidth} \text{(g)}
    \end{subfigure}
	\caption{Scaling the width of the input image by 50\%. (a) Input image, (b) Linear Scale, (c) Seam Carving~\cite{Rubinstein08}, (d) Warping~\cite{Wolf07}, (e) Multiop~\cite{Rubinstein09}, (f) SV~\cite{Krahenbuhl09} and (g) DNR (ours).}
	\label{fig:retarget_me_comparison_50}
\end{figure*}
\begin{figure*}
    \centering
    \begin{subfigure}{\linewidth}
    \centering
    \raisebox{1cm}{\rotatebox[origin=t]{90}{Input Image}}
    \adjincludegraphics[height=2.65cm]{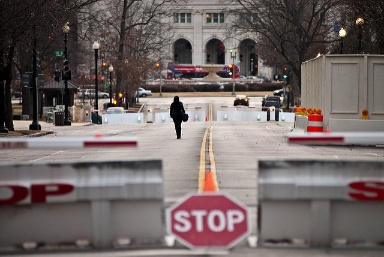}
    \adjincludegraphics[height=2.65cm]{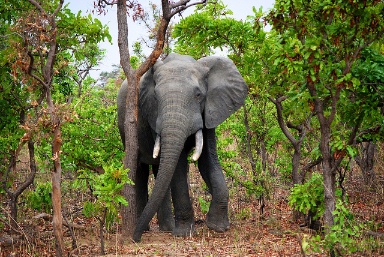}
    \adjincludegraphics[height=2.65cm]{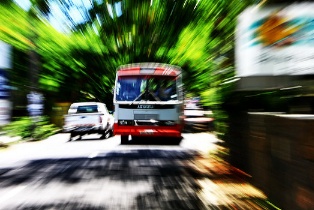}
    \adjincludegraphics[height=2.65cm]{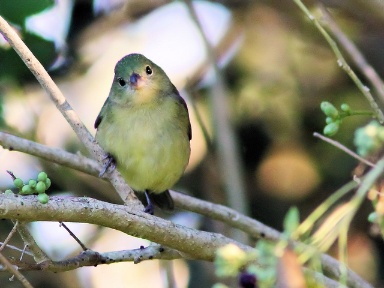}
    \end{subfigure}
    
    \begin{subfigure}{\linewidth}
    \centering
    \raisebox{0.75cm}{\rotatebox[origin=t]{90}{Linear Scale}}
    \hspace{0.00000001cm}
    \adjincludegraphics[height=2.1007895cm]{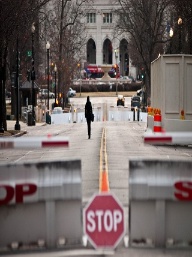}
    \adjincludegraphics[height=2.1007895cm]{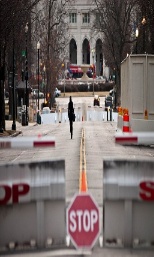}
    \adjincludegraphics[height=2.1007895cm]{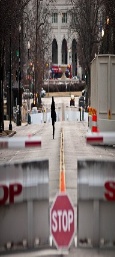}
    \adjincludegraphics[height=2.1007895cm]{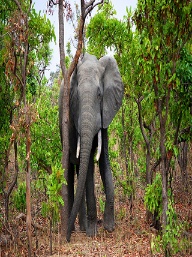}
    \adjincludegraphics[height=2.1007895cm]{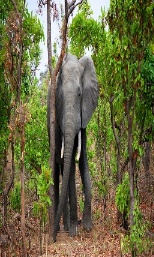}
    \adjincludegraphics[height=2.1007895cm]{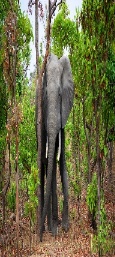}
    \adjincludegraphics[height=2.1007895cm]{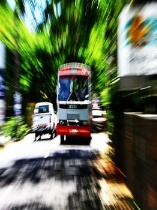}
    \adjincludegraphics[height=2.1007895cm]{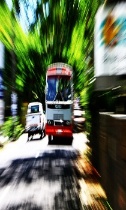}
    \adjincludegraphics[height=2.1007895cm]{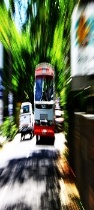}
    \adjincludegraphics[height=2.1007895cm]{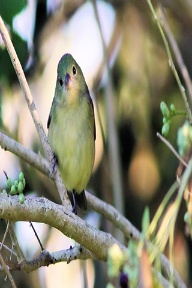}
    \adjincludegraphics[height=2.1007895cm]{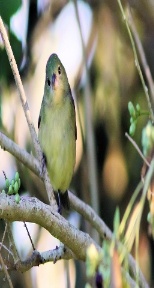}
    \adjincludegraphics[height=2.1007895cm]{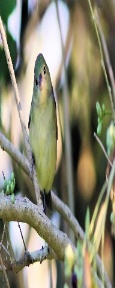}
    \end{subfigure}
    
    \begin{subfigure}{\linewidth}
    \centering
    \raisebox{0.7cm}{\rotatebox[origin=t]{90}{SC}}
    \hspace{0.00000001cm}
    \adjincludegraphics[height=2.1007895cm]{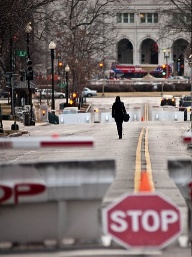}
    \adjincludegraphics[height=2.1007895cm]{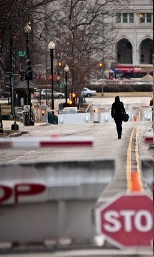}
    \adjincludegraphics[height=2.1007895cm]{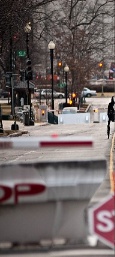}
    \adjincludegraphics[height=2.1007895cm]{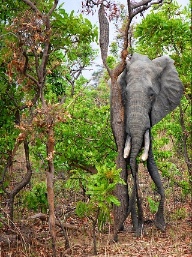}
    \adjincludegraphics[height=2.1007895cm]{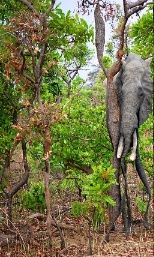}
    \adjincludegraphics[height=2.1007895cm]{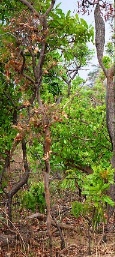}
    \adjincludegraphics[height=2.1007895cm]{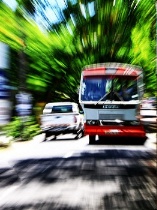}
    \adjincludegraphics[height=2.1007895cm]{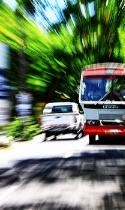}
    \adjincludegraphics[height=2.1007895cm]{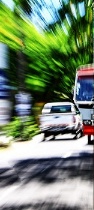}
    \adjincludegraphics[height=2.1007895cm]{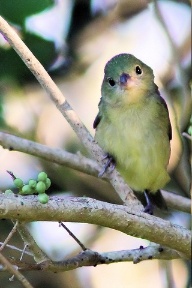}
    \adjincludegraphics[height=2.1007895cm]{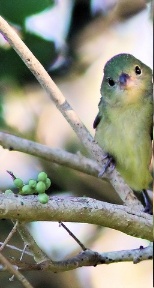}
    \adjincludegraphics[height=2.1007895cm]{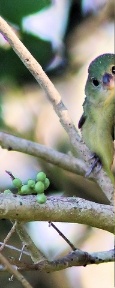}
    \end{subfigure}
    
    \begin{subfigure}{\linewidth}
    \centering
    \raisebox{0.75cm}{\rotatebox[origin=t]{90}{Ours}}
    \hspace{0.00000001cm}
    \adjincludegraphics[height=2.1007895cm]{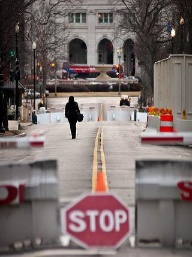}
    \adjincludegraphics[height=2.1007895cm]{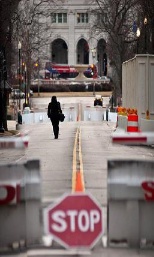}
    \adjincludegraphics[height=2.1007895cm]{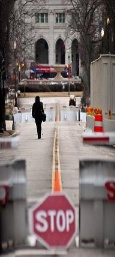}
    \adjincludegraphics[height=2.1007895cm]{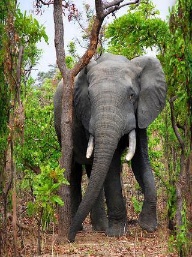}
    \adjincludegraphics[height=2.1007895cm]{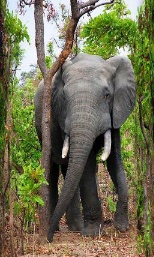}
    \adjincludegraphics[height=2.1007895cm]{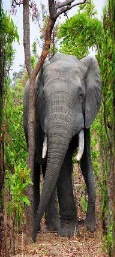}
    \adjincludegraphics[height=2.1007895cm]{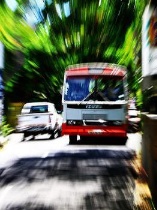}
    \adjincludegraphics[height=2.1007895cm]{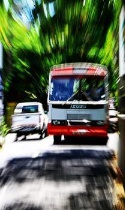}
    \adjincludegraphics[height=2.1007895cm]{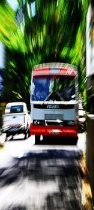}
    \adjincludegraphics[height=2.1007895cm]{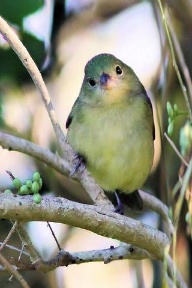}
    \adjincludegraphics[height=2.1007895cm]{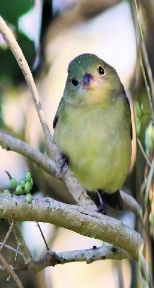}
    \adjincludegraphics[height=2.1007895cm]{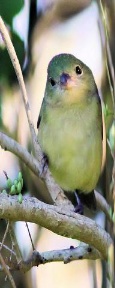}
    \end{subfigure}
    
    \caption{Stress test of extreme retargeting on images from Coco Dataset \cite{Coco}. The width of the input images (first row) is scaled by 50\%,40\% and 30\%. We compare linear scale (second row), seam carving~\cite{Rubinstein08} (third row) and our results (last row). Notice how the important subject in each image preserves its shape as much as possible.}
    \label{fig:stress_test}
\end{figure*}
We use the RetargetMe benchmark~\cite{Rubinstein10Comparative} containing a variety of images and the results of previous retargeting operators on these images.
We show sample results of DNR compared to Linear Scale, Seam Carving~\cite{Rubinstein08}, Warping~\cite{Wolf07} and MultiOp~\cite{Rubinstein09} in Figure~\ref{fig:retarget_me_comparison_75} and in Figure~\ref{fig:retarget_me_comparison_50}. As can be seen from both figures, DNR better retains the aspect ratios of semantic regions compared to the other methods. In addition, we compare DNR with the work in~\cite{Cho17} and~\cite{DBLP:journals/corr/abs-1811-07793}, and show the results in Figure~\ref{fig:ours_vs_wssdcnn} and Figure~\ref{fig:ours_vs_deepir}, respectively. We also demonstrate our method's ability in extreme size retargeting, and show more results in Figure~\ref{fig:stress_test}.

\subsection{User Study}

To evaluate our DNR method against other alternative methods we turned to RetargetMe benchmark~\cite{Rubinstein10Comparative} that compared various methods. We conducted two forced choice tests comparing our results side-by-side to an alternative. We showed the original image before retargeting and asked the user to choose the image that best preserves the content of the original image. The order of presentation was randomly shuffled and the survey forms were randomly distributed among 112 participants. 

First, we chose to compare against the best performing method, which is SV method~\cite{Krahenbuhl09}. DNR received 55.5\% of the votes when compared with SV (out of 889 votes in total). Second, we compared against the best result obtained per image. In other words, we chose the highest ranking results in the original study, where each image could be obtained using a different retargeting method. Even in this case, our results received 52.8\% of the votes (out of 956 votes in total).
Counting the number of images that users preferred our results we found that DNR was favored in 42 images (against 25 to SV), and in 37 images (against 29 to Best).

\subsection{Semantic Preservation Evaluation}
\label{QuantitativeEvaluation}
\begin{table}
    \begin{adjustbox}{width=\columnwidth,center}
		\begin{tabular}{ lccccccc }
			\toprule
			Method  & CR & SCL & SV & SC & WARP & Best & \textbf{DNR}   \\
			\midrule
			Avg. SS & 68\% & 68\% & 64\% & 68\% & 65\% & 65\% & \textbf{70\%}  \\
			\bottomrule
		\end{tabular}
    \end{adjustbox}
	\caption{Average Semantic Score on RetargetMe. The comparison is made between Manual Crop (CR), Linear Scale (SCL), Streaming Video (SV\cite{Krahenbuhl09}), Seam Carving (SC\cite{Rubinstein08}) and Warpping\cite{Wolf07}. Further, we include the average semantic score computed on the best retargeted images that were chosen in RetargetMe user study (Best).
	}
	\label{tbl:semantic_score}
\end{table}

We want to compare the preservation of semantic details as a result of the retargeting operator. For this purpose, we define a \textbf{Semantic Score (SS)} given by:

\begin{equation}\label{eq:SemanticScore}
\begin{split}
\mathcal{SS}_{i} & = \frac{ \norm{\F_{i}(\mathcal{O})}_{2} } {\norm{ \F_{i}(\I)}_{2} }
\end{split}   
\end{equation} 

This score compares the magnitudes of some deep VGG-19 layer activation, before and after the retargeting. In particular, we expect that if the retargeting operator damages semantic regions, then the score will be lower, since in this case high activation on the original image will increase the denominator $\norm{\F_{i}(\mathcal{I})}_2$, while low activation on the retrageted image will diminish the numerator $\norm{\F_{L}(\mathcal{O})}_2$. We used $block_5\_conv_1$ as the feature map $\F_i(\cdot)$ in Equation~\ref{eq:SemanticScore}.
In Table~\ref{tbl:semantic_score}, we computed the average semantic score per image  in the RetargetMe benchmark for different retargeting operators. As can be seen, our DNR method receives the highest score.

\subsection{Limitations}\label{Limitations}
\begin{figure}
    \centering
    \includegraphics[width=8cm]{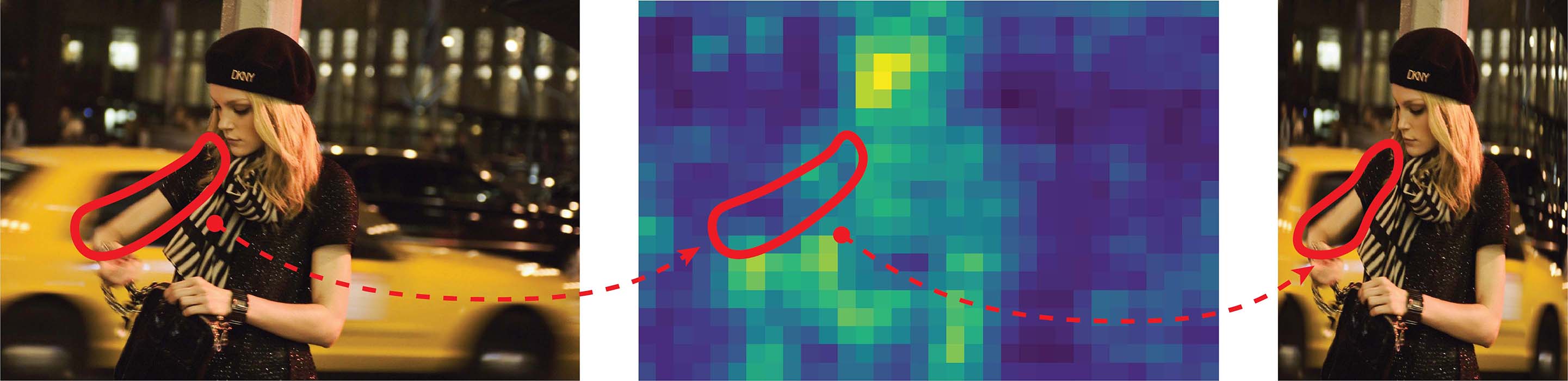}
    \caption{When the neural network does not recognize important parts on an image (left), the corresponding deep layer has low activation (middle). In this case the hand of the woman is not detected and this leads to unwanted distortion in the output image (right).}
    \label{fig:limitations_a}
\end{figure}

\begin{figure}
    \centering
    \vspace{0.25cm}
    \includegraphics[width=8cm]{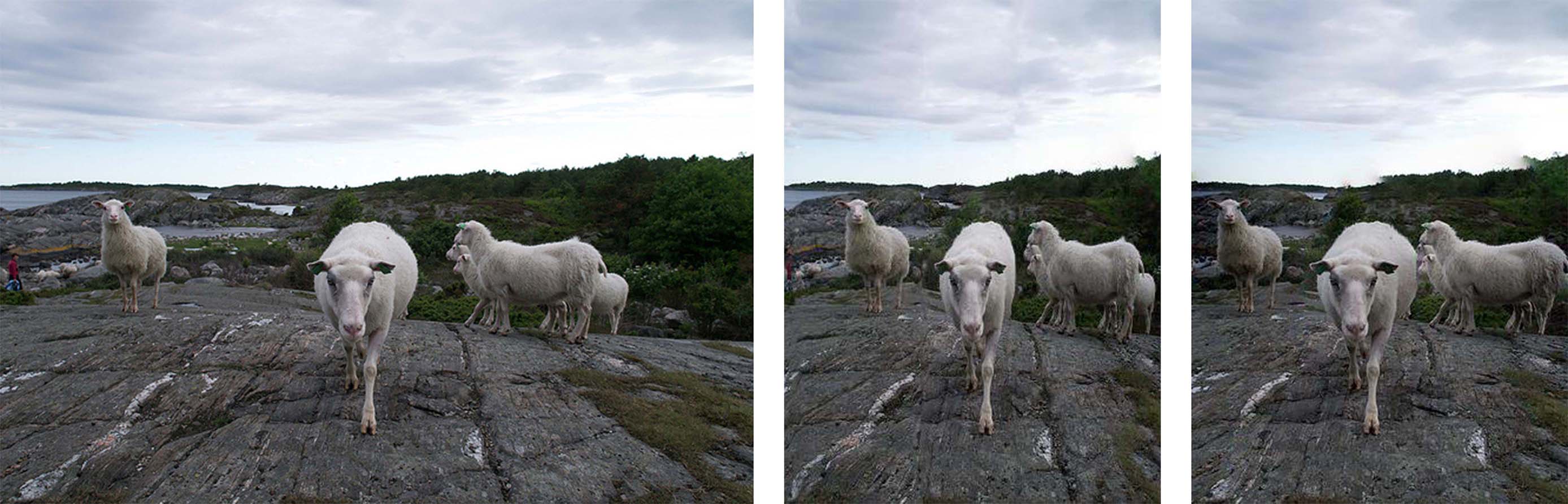}
    \caption{Changing the threshold determining when to switch from seam carving to warping can lead to different results. In this example, our automatic results (middle) does not preserve the aspect ratio of the sheep compared to applying only Deep Seam Carving (right).}
    \label{fig:limitations_b}
\end{figure}

In our experiments, we encountered two challenges that resulted in unsatisfactory results. 

First, VGG19~\cite{Simonyan14} was trained for the purpose of object detection, and DNR relies on its ability to detect semantic regions and objects. However, the network doesn't always succeed in providing semantic information on important regions. In addition, the network detects specific features in an object and can still have low activation on different regions of an important objects. All these could lead to object distortion in the final results (see Figure~\ref{fig:limitations_a}).

Another challenge is choosing the correct threshold to switch from seam-carving to warping in our multi-operator scheme. In particular, we have seen that there are cases in which further seams could have been removed while in other cases, we removed too many seams (see Figure~\ref{fig:limitations_b}). 

Lastly, the time to produce results using DNR is still large. On average, it takes up to two minutes to retarget an image of size $640\times480$. Further optimization is required to achieve online image retargeting which can lead to possible extension of DNR to videos.
\section{Conclusion}
We have presented an image retargeting technique that operates in deep layers of a pre-trained neural network. The technique utilizes the semantic information latent in the deep layers hierarchy to aggregate on-the-fly an effective importance map. We have shown the strength of the high-level image analysis versus common low-level feature analysis. In addition, our technique is based on an optimization procedure that reconstructs the image from its deep features, hence, it tends to produce much less visual artifacts. 

In this work, we use a specific available pre-trained network. However, in the future we would like to consider pre-training a network with a special-purpose target in mind, so its deep features will be more relevant to a specific task. 
Another avenue for future work is to leverage the optimization of the target image to synthesize new content. This will possibly be effective in upscaling an image into an overly different aspect ratio. 

\section*{Acknowledgment}
This research was supported by Israel Science Foundation Grant number 1390/19.



{\small
\bibliographystyle{ieee}
\bibliography{egbib}
}

\end{document}